\documentclass{article} 
\usepackage{iclr2026_conference,times}

\usepackage{amsmath,amsfonts,bm}

\def\eqref#1{equation~\ref{#1}}

\def\1{\bm{1}}

\DeclareMathAlphabet{\mathsfit}{\encodingdefault}{\sfdefault}{m}{sl}
\SetMathAlphabet{\mathsfit}{bold}{\encodingdefault}{\sfdefault}{bx}{n}

\usepackage{subcaption}
\usepackage{hyperref}
\usepackage{url}
\usepackage{booktabs}
\usepackage[utf8]{inputenc} 
\usepackage[T1]{fontenc}    
\usepackage{hyperref}       
\usepackage{url}            
\usepackage{booktabs}       
\usepackage{amsfonts}       
\usepackage{nicefrac}       
\usepackage{microtype}      
\usepackage{xcolor}         
\usepackage{soul}
\usepackage{graphicx}
\usepackage{amsmath}
\usepackage{multirow}
\usepackage{tabularx}
\usepackage{colortbl}
\usepackage{wrapfig} 
\usepackage{pifont}
\usepackage{comment}
\AtBeginEnvironment{table}{\tiny} 
\AtBeginEnvironment{wraptable}{\tiny} 

\title{Autoencoding-Free Context Compression for LLMs via Contextual Semantic Anchors}

\author{%
Xin Liu$^{1,}$\thanks{Equal contribution. Email: \texttt{\{liuxin1,zhaors\}@mails.neu.edu.cn}}
\quad
Runsong Zhao$^{1,*}$
\quad
Pengcheng Huang$^{1}$ 
\quad
Xinyu Liu$^{1}$ 
\quad
Junyi Xiao$^{1}$ \\
\textbf{Chunyang Xiao
\quad
Tong Xiao$^{1,2,}$\thanks{Corresponding author. Email: \texttt{xiaotong@mail.neu.edu.cn}}  
\quad
Shengxiang Gao$^{3}$ 
\quad
Zhengtao Yu$^{3}$ 
\quad
Jingbo Zhu$^{1,2}$} \\
  $^1$ School of Computer Science and Engineering, Northeastern University, Shenyang, China \\
  $^2$ NiuTrans Research, Shenyang, China \\
  $^3$ Kunming University of Science and Technology, Kunming, China \\
}

\iclrfinalcopy 
\usepackage{float}
\begin{document}
\maketitle


\begin{abstract}
Context compression is an advanced technique that accelerates large language model (LLM) inference by converting long inputs into compact representations.
Existing methods primarily rely on autoencoding tasks to train special compression tokens to represent contextual semantics.
While autoencoding tasks enable compression tokens to acquire compression capabilities, we remark that such capabilities potentially conflict with actual downstream
task requirements, prevent the models from learning the features more beneficial for real-world usage.
Based on this observation, we propose Semantic-Anchor Compression (SAC), a novel method that shifts from autoencoding task based compression to an architecture that is equipped with this compression capability \textit{a priori}. Instead of training models to compress contexts through autoencoding tasks, SAC directly selects so-called anchor tokens from the original context and aggregates contextual information into their key-value (KV) representations. To ensure that anchors can effectively collect information, SAC introduces two key designs: (1) anchor embedding, a learnable embedding vector attached to the selected anchor tokens to mark compression carriers and (2) bidirectional attention modification, which enables anchor tokens to integrate information from the entire context. Experimental results show that SAC consistently outperforms existing context compression methods across different compression ratios and model sizes on question-answering and long-context summarization tasks. Our data, model and code have been released at \href{https://github.com/lx-Meteors/SAC}{https://github.com/lx-Meteors/SAC}.
\end{abstract}

\section{Introduction}

The expanding scope of large language models (LLMs) to tasks such as processing long documents~\citep{liu2024forgetting,li-etal-2024-loogle,duan2025chunks,zhao2026cometcollaborativememorytransformer}, maintaining multi-turn dialogue coherence~\citep{zhang2025surveymultiturninteractioncapabilities,yi2025surveyrecentadvancesllmbased,guan2025evaluatingllmbasedagentsmultiturn}, and generating responses grounded in extensive external knowledge~\citep{10.5555/3495724.3496517,karpukhin-etal-2020-dense,huang2025pip} necessitates the incorporation of vast contexts into the model input. However, directly processing such extremely long contexts is fraught with challenges, including prohibitive computational costs, significant inference latency, and performance degradation, largely caused by the ``lost-in-the-middle'' phenomenon~\citep{liu-etal-2024-lost}.

\begin{figure*}[tbp]
    \centering
    \includegraphics[width=\textwidth]{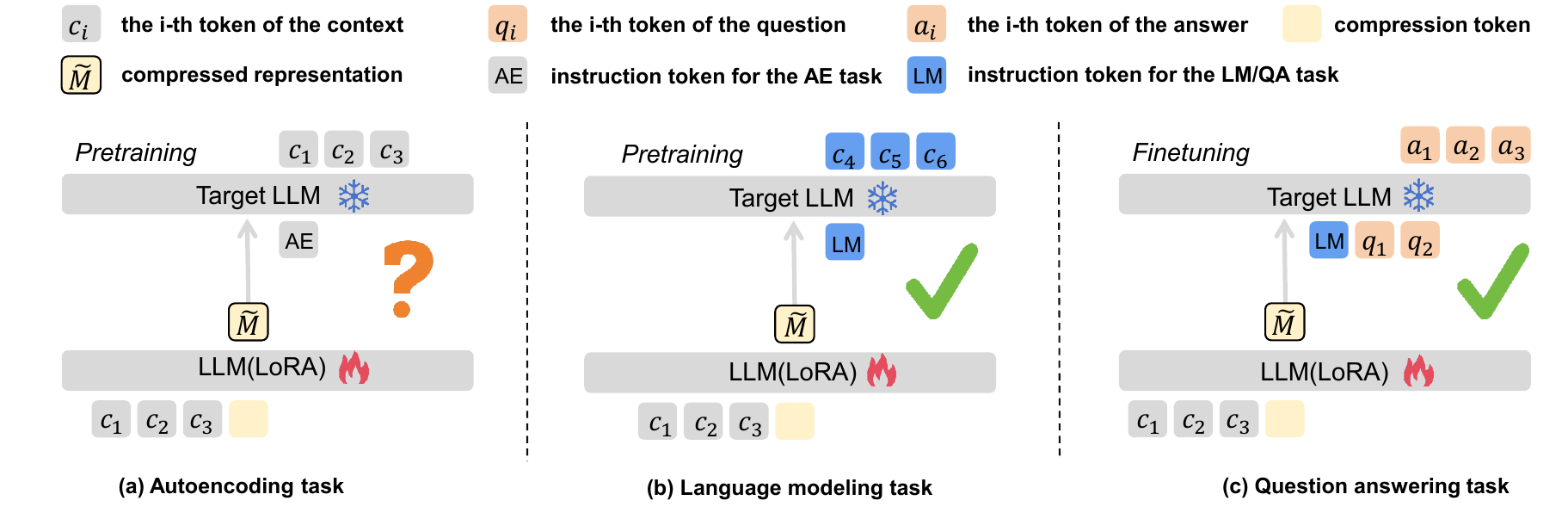}
    \caption{Three tasks for training the context compressor introduced by ICAE and followed by numerous works. The training uses (a) Autoencoding task and (b) Language modeling task to pretrain the encoder, then finetunes on (c) Question answering task.}
    \label{fig:ae_vs_lm_task}
\end{figure*}

To address these challenges, recent studies have proposed context compression~\citep{chang2024efficientpromptingmethodslarge,li-etal-2025-prompt,lv2026datadistributionmattersdatacentric,tang2026readhumancompressingcontext,tang2026comicoarsetofinecontextcompression}, a technique that typically appends special tokens (i.e. compression tokens) to the end of the context and leverages the LLM's causal attention mechanism to compress contextual information into a compact representation within these tokens. Once this compact representation is obtained, the LLM can generate responses conditioned on it, rather than being conditioned on the entire original context. The reduction in context length leads to substantial decreases in both inference time and GPU memory consumption. While effective, these approaches face a key limitation: the compression tokens are randomly initialized and lack inherent semantic information. To address this limitation, prior works~\citep{ge2024incontextautoencodercontextcompression, wang2024incontextformerlightningfastcompressing, li2024500xcompressorgeneralizedpromptcompression, zhao2025positionidsmatterenhanced, tang2025gmsaenhancingcontextcompression} typically rely on extensive pretraining on autoencoding (AE) and language modeling (LM) tasks (as shown in Fig.~\ref{fig:ae_vs_lm_task}) to endow compression tokens with the ability to carry contextual information. However, while the model learns to reconstruct contexts from compression tokens through the AE task, such reconstruction objective can misalign with actual objectives (e.g. QA from compressd tokens) that downstream requires (we analyse such misalignment in subsection~\ref{subsec:ae_analysis}). Thus, this reliance on a suboptimal and costly pretraining stage raises a critical research question: is it possible to design a compression architecture that inherently understands context without a demanding AE phase?

To answer this question, our work introduces Semantic-Anchor Compression (SAC) (Figure~\ref{fig:SAC}), a novel architecture for the context compression task. Instead of appending new special tokens and requiring extensive autoencoding pretraining to learn their representations, SAC directly selects representative tokens from the original context to act as ``anchor tokens'' for compression. By leveraging these semantically meaningful anchors from the input itself, SAC incorporates natural semantic priors that obviate the need for autoencoding  pretraining.
To mark their special role, these selected tokens are augmented with dedicated ``anchor embeddings'', enabling the LLM to distinguish them from regular tokens. Furthermore, to enhance their compression capabilities, we modify the standard causal attention to a bidirectional attention mechanism. This allows anchor tokens to access information from the entire context, rather than being restricted to only preceding tokens. These modifications collectively foster a more effective context compression by providing anchor tokens with both distinct representations and comprehensive contextual awareness. To evaluate the effectiveness of SAC, we conduct experiments on question answering tasks and long-context summarization tasks, where it consistently outperforms strong baselines. 
Results also show that 1) our proposed method consistently improves over strong baselines across different compress ratios, 2) our proposed architecture achieves its best performance in a simpler training setting without autoencoding training, arguably because the anchor tokens already contain enough information about the original context.
Our analysis reveals that SAC's compressed representations more closely resemble original context token KVs in feature space, which arguably enables LLMs to better understand them during inference.

\begin{figure*}[htbp]
    \centering
    \includegraphics[width=\textwidth]{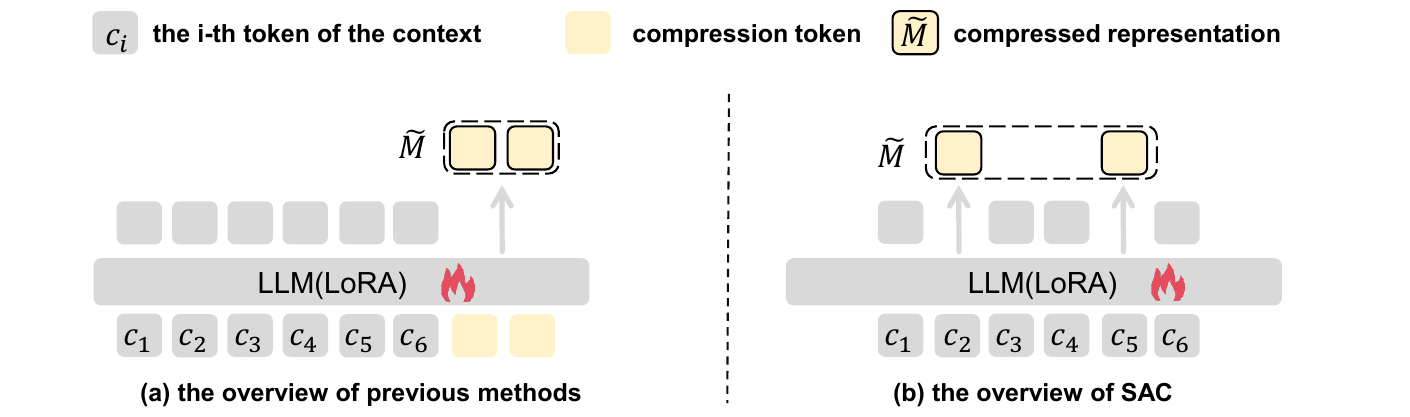}
     \caption{The difference between SAC and previous methods. While previous methods (a) compress contextual information into dedicated special tokens (referred to as compression tokens), SAC (b) compresses the context directly into the original contextual tokens themselves. Here, $\tilde{M}$ can represent either the output from the final layer of the LLM or the key-value pairs across all transformer layers, which are later used as compressed representations for LLM inference.}
\label{fig:SAC}
\end{figure*}

\section{Related Works}

\subsection{Compression method}

Many methods focus on reducing prompt lengths to improve LLM inference speed. CC~\citep{wingate-etal-2022-prompt} utilizes contrastive learning to compress specific natural language prompts into shorter and unique soft prompt tokens. However, it cannot generalize to unseen prompts and requires retraining for new prompts. GIST~\citep{mu2023learning} compresses original prompts into KV values through finetuning and can handle arbitrary unseen contexts. AutoCompressor~\citep{chevalier-etal-2023-adapting} recursively combines compressed vectors with sub-prompts and aggregates all compressed vectors to construct the final representation, enabling compression of longer contexts. However, both GIST and AutoCompressor require finetuning the LLMs (referred to later as target LLMs) to perform inference, which may affect LLMs' original capabilities.

ICAE~\citep{ge2024incontextautoencodercontextcompression} formulates context compression as training a general encoder that compresses contexts into compact representations understandable by target LLMs without finetuning target LLMs. To train the encoder, ICAE introduces autoencoding tasks and performs joint pretraining with language modeling tasks, followed by finetuning on downstream tasks. 500xCompressor~\citep{li2024500xcompressorgeneralizedpromptcompression} improves upon ICAE by replacing the compression carrier from the last layer output of compression tokens with KV values at each layer, achieving higher compression ratios. EPL~\citep{zhao2025positionidsmatterenhanced} identifies that ICAE and 500xCompressor neglect the impact of positional encoding and proposes distributing compression token position IDs uniformly across the entire context rather than placing them at the end. However, these methods still rely on autoencoding tasks to endow the compression tokens with the ability to carry contextual information.

Another category of prompt compression methods is based on token selection, which selects representative tokens from contexts based on token importance. SelectiveContext~\citep{li-etal-2023-compressing}, LLMLingua~\citep{jiang-etal-2023-llmlingua}, and LongLLMLingua~\citep{jiang-etal-2024-longllmlingua} employ causal small language models to evaluate token importance based on information entropy. LLMLingua-2~\citep{pan-etal-2024-llmlingua} distills a token classifier to compute the probability of each token to be preserved. PerceptionCompressor~\citep{tang-etal-2025-perception} preserves critical information at the token level while removing tokens that distract the LLM’s attention. These works demonstrate that LLMs can understand original contexts using a small number of representative tokens. However, they do not perform compressed tokens training which limits the usability of the selected tokens by target LLMs. Our proposed SAC can be seen as a combination of token selection methods and compressed token training methods: it derives and train compressed representations that are based on tokens selected directly from the context and indeed is compatible with the token selection methods above.

\subsection{Bidirectional Attention}
Recent studies have shown that, removing the decoder's unidirectional causal constraint and introducing bidirectional attention can effectively enhance the model's representational capacity~\citep{wang2020supergluestickierbenchmarkgeneralpurpose}. For instance, NV-Embed~\citep{lee2025nvembedimprovedtechniquestraining} replaces  causal attention with bidirectional attention during contrastive training, achieving strong performance on general text embedding and dense vector retrieval tasks. LLM2Vec~\citep{behnamghader2024llm2veclargelanguagemodels}, by enabling bidirectional attention alongside masked next-token prediction, significantly improves the model's ability to capture global semantics in text embedding tasks. These works indicate that bidirectional attention is advantageous for acquiring global semantic information and robust contextual representations. However, its effectiveness in context compression tasks remains underexplored. Motivated by these findings, we incorporate bidirectional attention into the compressor to enhance contextual modeling during the compression phase.

\section{Method}
\label{headings}
\subsection{Task Formulation}

Context compression is formally defined as follows: an encoder $\mathcal{E}$ compresses a context $C = (c_1, c_2, \ldots, c_{|C|})$ into a compact representation $\tilde{M}$ with $\tilde{M} = \mathcal{E}(C)$.
Subsequently, a target LLM leverages the compressed representation $\tilde{M}$ in place of the original context $C$ to perform various tasks, such as question answering.

To train the encoder $\mathcal{E}$ to effectively extract contextual information, ICAE introduces three objective functions.
The autoencoding loss $\mathcal{L}_{\mathbf{AE}}$ ensures that the compressed representation $\tilde{M}$ generated by $\mathcal{E}$ preserves all tokens in the context by learning to reconstruct the entire context $C$ regardless of the token relative importance, as shown in Figure~\ref{fig:ae_vs_lm_task}a; mathematically, $\mathcal{L}_{\mathbf{AE}} = -\log P(C|\tilde{M})$. The language modeling loss $\mathcal{L}_{\mathbf{LM}}$ encourages $\tilde{M}$ to maintain predictive capability for future context $C'=(c_{|C|+1}, c_{|C|+2}, \ldots, c_{|C|+|C'|})$, enabling proactive information planning, as shown in Figure~\ref{fig:ae_vs_lm_task}b; mathematically, $\mathcal{L}_{\mathbf{LM}} = -\log P(C'|\tilde{M})$. During pretraining, $\mathcal{L}_{\mathbf{AE}}$ and $\mathcal{L}_{\mathbf{LM}}$ are jointly optimized to obtain an initially effective encoder $\mathcal{E}$.

Additionally, during finetuning, the downstream task loss $\mathcal{L}_{\mathbf{QA}}$ (taking QA as an exemplar downstream task) enhances the ability of $\tilde{M}$ to extract information that is potentially relevant for downstream tasks. For example, for question answering (QA) tasks, the encoder learns through $\mathcal{L}_{\mathbf{QA}}$ to identify and preserve information in the context that is likely to be queried, enabling accurate answer generation $A = (a_1, a_2, \ldots, a_{|A|})$ when presented with subsequent questions $Q = (q_1, q_2, \ldots, q_{|Q|})$, as shown in Figure~\ref{fig:ae_vs_lm_task}c; mathematically, $\mathcal{L}_{\mathbf{QA}} = -\log P(A|\tilde{M},Q)$.


\subsection{Semantic-Anchor Compressor}

\begin{figure*}[htbp]
    \centering
    \includegraphics[width=\textwidth]{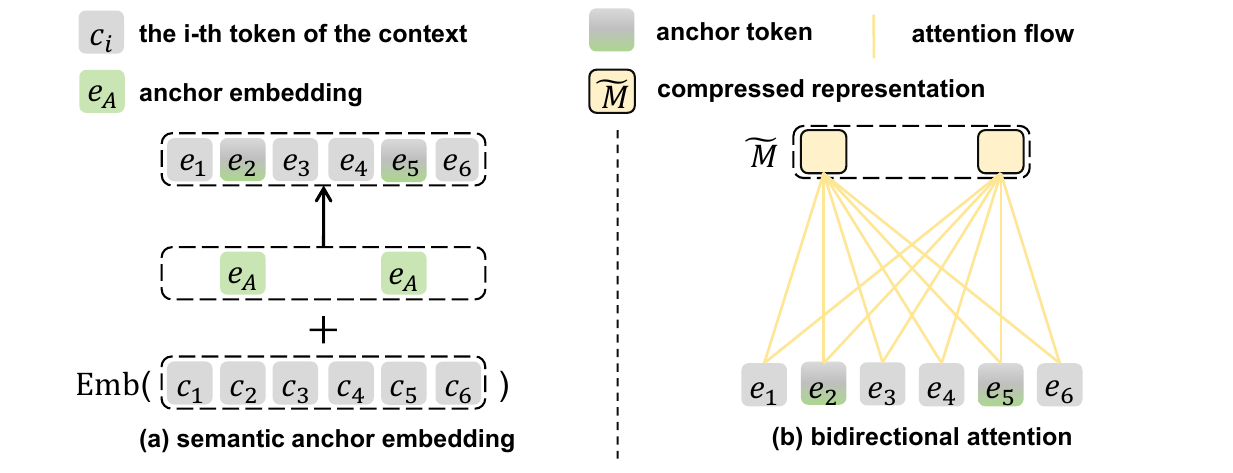}
     \caption{Key differentiators within SAC model architecture. (a) Representative tokens are transformed into anchor tokens through anchor embeddings. (b) The encoder in SAC adopts bidirectional attention, while the decoder operates with causal attention. 
}
    \label{fig:attention-anchor}
\end{figure*}

A key distinction between our approach Semantic-Anchor Compression (SAC) and previous methods is that we derive compressed representations directly from selected context tokens, as shown in Figure~\ref{fig:SAC}. This involves selecting a subset of tokens $S$ from context $C$ as anchor tokens $S \subseteq C$. We believe that a good selection strategy benefits SAC. Following EPL, our default strategy divides the entire context $C$ into $|S|$ chunks and selects the middle token from each chunk. This setting helps maximize coverage of context $C$.
As illustrated in Figure~\ref{fig:attention-anchor}a, each selected token $c_i \in S$ is enhanced with the anchor embedding $e_A$, yielding an embedding sequence $E=(e_1, e_2, \ldots, e_{|C|})$: \begin{equation} e_i = \mathbf{Emb}(c_i) + \mathbf{1}_{c_i \in S} \cdot e_A \end{equation} where $\mathbf{1}_{c_i \in S}$ is an indicator function that equals 1 when $c_i \in S$ and 0 otherwise. Following previous works, we employ a LLM with LoRA parameters $\theta_{LoRA}$ as the compressor: $\tilde{M} = \mathcal{E}(C) = \mathbf{LLM}(E|\theta_{LoRA}) $. Overall, using original tokens from the context avoids learning compression tokens from scratch and we empirically validate that such design improves learning efficiency. 

We notice that because the encoder uses causal attention, the anchor tokens $S$ do not have visibility to the full sentence, limiting their representation power. Hence in SAC, we modify the LLM to replace causal attention with bidirectional attention (see Figure~\ref{fig:attention-anchor}b), to enhance the LLM's encoding capability. Notably, this bidirectional attention operates across all tokens, rather than being limited to the anchor tokens, allowing the model to capture richer contextual dependencies. 

$\tilde{M}$ can be either the output of anchor tokens from the LLM's final layer or the key-value pairs from each layer. Following 500xCompressor, we use key-value pairs as the compressed representation $\tilde{M}$. For the encoder, we adopt the same language model as the decoder and leverage the original context's KV cache, enabling the decoder to comprehend the compressed representation more readily without requiring additional semantic alignment.\footnote{For context compression, adopting different LLMs for encoder and decoder will create semantic gaps between the two and will impair performance, as noticed by \citep{lv2026datadistributionmattersdatacentric}.} During pretraining, we only use $\mathcal{L}_{\mathbf{LM}}$ and do not use $\mathcal{L}_{\mathbf{AE}}$ to train the compressor. Following previous work, we use $\mathcal{L}_{\mathbf{QA}}$ for finetuning.

\section{Experiments}

\subsection{Experimental Setting}

\textbf{Dataset.} For continued pretraining, we use the corpus~\href{https://huggingface.co/datasets/DKYoon/SlimPajama-6B}{SlimPajama-6B}~\citep{cerebras2023slimpajama}. During finetuning and evaluation, we employ the standard \href{https://huggingface.co/datasets/mrqa-workshop/mrqa}{MRQA}~\citep{fisch2019mrqa} question-answering dataset, which consolidates multiple QA tasks and standardizes them into a unified format. We evaluate SAC on both test sets, namely in-domain (ID) and out-of-domain (OOD), to comprehensively assess its in-distribution fitting ability and cross-domain generalization performance.

\textbf{Implementation Details.} SAC utilizes \href{https://huggingface.co/meta-llama/Llama-3.2-1B}{Llama-3.2-1B}~\citep{grattafiori2024llama3herdmodels} as both the encoder and target LLM. 
We equip the encoder with trainable LoRA~\citep{hu2022lora} adapters (rank = 128, $\alpha = 256$), while the target LLM parameters remain frozen. For each context, we partition it into sub-contexts of 510 tokens each. The compressor compresses each sub-context into a sub-compressed representation, and subsequently concatenates these sub-compressed representations to form the complete compressed representation. The number of anchor tokens $|S| = \lfloor L / r \rfloor$ is determined by the compression ratio $r$ and the length of the sub-context $L$. We train all models in two stages: pretraining for 20,000 optimization steps followed by finetuning for an additional 20,000 steps, both conducted with a batch size of 16. Complete hyperparameter configurations are provided in Appendix~\ref{app:exp}.

\textbf{Baselines.} We use the Llama-3.2-1B model trained on the MRQA~\citep{fisch2019mrqa} dataset as an uncompressed baseline (denoted as ``Full-FT"). We compare our method against several context compression techniques: for hard compression, we choose LLMLingua-2 \citep{pan2024llmlingua2datadistillationefficient} and evaluate its performance on the Full-FT model; for soft compression, we select ICAE~\citep{ge2024incontextautoencodercontextcompression}, 500xCompressor \citep{li2024500xcompressorgeneralizedpromptcompression}, Activation Beacon~\citep{zhang2024longcontextcompressionactivation}, DAST~\citep{chen2025dastcontextawarecompressionllms}, and EPL \citep{zhao2025positionidsmatterenhanced}. To ensure a fair comparison, all these soft compression baselines are trained using the same experimental settings and data as our SAC method.

\begin{table*}[ht!]
    \small
    \centering
    \caption{For the fine-tuning results, we report in-domain performance on the MRQA datasets using ROUGE-1 F1~\citep{lin-2004-rouge} and exact match (EM)~\citep{maalouly2022exactmatchingalgorithmsrelated}.The compression ratio of Lingua-2~\citep{pan2024llmlingua2datadistillationefficient} is set to 5×.
}
    \label{tab:sft-iid-results}
    \resizebox{\textwidth}{!}{
    \begin{tabular}{l|cccccccccccccc}
        \toprule
        \multirow{2}{*}{\textbf{Methods}} 
        & \multicolumn{2}{c}{\textbf{SQuAD}} 
        & \multicolumn{2}{c}{\textbf{NewsQA}} 
        & \multicolumn{2}{c}{\textbf{TriviaQA}} 
        & \multicolumn{2}{c}{\textbf{SearchQA}}  
        & \multicolumn{2}{c}{\textbf{HotpotQA}} 
        & \multicolumn{2}{c}{\textbf{NQ}}  
        & \multicolumn{2}{c}{\textbf{Average}}\\
        \cmidrule(r){2-3} \cmidrule(r){4-5} \cmidrule(r){6-7} 
        \cmidrule(r){8-9} \cmidrule(r){10-11} \cmidrule(r){12-13} \cmidrule(r){14-15}
        & \textbf{F1} & \textbf{EM} 
        & \textbf{F1} & \textbf{EM} 
        & \textbf{F1} & \textbf{EM} 
        & \textbf{F1} & \textbf{EM} 
        & \textbf{F1} & \textbf{EM} 
        & \textbf{F1} & \textbf{EM} 
        & \textbf{F1} & \textbf{EM}\\ 
        
        \midrule
        Full-FT &77.69&59.71 &63.50&46.04 &68.80&60.54 &73.25&62.07 &74.78&59.26 &71.01&53.47 &71.51&56.85  \\
        Lingua-2 &32.93&19.57 &26.78&13.20 &9.67&8.12 &45.40&31.80 &36.10&22.05 &40.08&22.01 &31.83&19.46  \\
        \midrule
        \multicolumn{14}{c}{{\textit{15x compression constraint}}} \\
        \midrule
        ICAE &31.90&18.91 &25.25&11.97 &51.78&42.94 &64.81&52.89 &45.22&30.32 &48.01&30.67 &44.50&31.28 \\
        500x &40.68&24.97 &32.01&16.76 &53.84&44.86 &65.65&53.70 &53.01&36.30 &50.93&33.26 &49.35&34.98 \\
        Beacon &{37.53}&{14.31} &{31.10}&{7.72} &{48.85}&{4.47} &{39.06}&{22.67} &{45.01}&{27.35} &{40.29}&{15.82} &{40.31}&{15.39} \\
        DAST &{36.33}&{22.65} &{31.55}&{16.29} &{56.92}&{48.14} &{68.07}&{56.51} &{54.02}&{37.93} &{52.10}&{35.31} &{49.83}&{36.14} \\
        EPL &44.58&27.91 &33.34&16.69 &56.16&47.09 &66.36&54.13 &54.88&38.38 &53.80&35.71 &51.52&36.65 \\
        {\cellcolor[rgb]{0.925,0.957,1}}\textbf{SAC} 
        & {\cellcolor[rgb]{0.925,0.957,1}}\textbf{47.43} & {\cellcolor[rgb]{0.925,0.957,1}}\textbf{30.25} 
        & {\cellcolor[rgb]{0.925,0.957,1}}\textbf{36.55} & {\cellcolor[rgb]{0.925,0.957,1}}\textbf{18.07} 
        & {\cellcolor[rgb]{0.925,0.957,1}}\textbf{61.13} & {\cellcolor[rgb]{0.925,0.957,1}}\textbf{52.19} 
        & {\cellcolor[rgb]{0.925,0.957,1}}\textbf{68.97} & {\cellcolor[rgb]{0.925,0.957,1}}\textbf{56.76} 
        & {\cellcolor[rgb]{0.925,0.957,1}}\textbf{58.83} & {\cellcolor[rgb]{0.925,0.957,1}}\textbf{41.86} 
        & {\cellcolor[rgb]{0.925,0.957,1}}\textbf{56.79} & {\cellcolor[rgb]{0.925,0.957,1}}\textbf{38.88} 
        & {\cellcolor[rgb]{0.925,0.957,1}}\textbf{54.95} & {\cellcolor[rgb]{0.925,0.957,1}}\textbf{39.67} \\
        \bottomrule
    \end{tabular}
    }
\end{table*}

\begin{table*}[ht!]
    \small
    \centering
    \caption{For the fine-tuning results, we report out-of-domain performance on the MRQA datasets using ROUGE-1 F1~\citep{lin-2004-rouge} and exact match (EM)~\citep{maalouly2022exactmatchingalgorithmsrelated}.The compression ratio of Lingua-2~\citep{pan2024llmlingua2datadistillationefficient} is set to 5×.
}
    \label{tab:sft-ood-results}
    \resizebox{\textwidth}{!}{
    \begin{tabular}{l|cccccccccccccc}
        \toprule
        \multirow{2}{*}{\textbf{Methods}} 
        & \multicolumn{2}{c}{\textbf{BioASQ}} 
        & \multicolumn{2}{c}{\textbf{DROP}} 
        & \multicolumn{2}{c}{\textbf{DouRC}} 
        & \multicolumn{2}{c}{\textbf{RACE}}  
        & \multicolumn{2}{c}{\textbf{RE}} 
        & \multicolumn{2}{c}{\textbf{TQA}}  
        & \multicolumn{2}{c}{\textbf{Average}}\\
        \cmidrule(r){2-3} \cmidrule(r){4-5} \cmidrule(r){6-7} 
        \cmidrule(r){8-9} \cmidrule(r){10-11} \cmidrule(r){12-13} \cmidrule(r){14-15}
        & \textbf{F1} & \textbf{EM} 
        & \textbf{F1} & \textbf{EM} 
        & \textbf{F1} & \textbf{EM} 
        & \textbf{F1} & \textbf{EM} 
        & \textbf{F1} & \textbf{EM} 
        & \textbf{F1} & \textbf{EM} 
        & \textbf{F1} & \textbf{EM}\\ 
    
        \midrule
        Full-FT &49.37&36.77 &44.67&34.46 &48.82&35.51 &35.57&9.64 &83.34&72.46 &53.32&32.40 &52.51&36.87  \\
        Lingua-2 &27.76&19.48 &27.28&18.83 &27.07&18.32 &17.54&4.15 &39.30&20.59 &28.42&15.83 &27.90&16.20  \\
        \midrule
        \multicolumn{14}{c}{{\textit{15x compression constraint}}} \\
        \midrule
        ICAE &35.51&24.47 &30.39&21.96 &13.78&9.06 &15.21&3.71 &55.24&40.33 &34.75&21.56 &30.81&20.18 \\
        500x &36.30&25.93 &33.46&23.55 &20.53&12.72 &18.49&3.41 &54.37&41.11 &41.09&25.82 &34.04&22.09 \\
        Beacon &{36.07}&{7.65} &{34.39}&{14.90} &\textbf{33.78}&{14.99} &\textbf{26.68}&{3.71} &{53.28}&{23.91} &{37.27}&{12.97} &{36.91}&{13.02} \\
        DAST &{36.57}&{27.06} &{31.90}&{22.42} &{21.56}&{14.26} &{16.31}&{3.26} &{48.54}&{35.92} &{36.69}&{22.75} &{31.93}&{20.95} \\
        EPL &40.52&28.52 &32.16&22.29 &25.70&16.39 &20.97&4.01 &59.75&46.34 &41.31&25.42 &36.74&23.83 \\
        {\cellcolor[rgb]{0.925,0.957,1}}\textbf{SAC} 
        & {\cellcolor[rgb]{0.925,0.957,1}}\textbf{41.31} & {\cellcolor[rgb]{0.925,0.957,1}}\textbf{28.66} 
        & {\cellcolor[rgb]{0.925,0.957,1}}\textbf{36.72} & {\cellcolor[rgb]{0.925,0.957,1}}\textbf{27.48} 
        & {\cellcolor[rgb]{0.925,0.957,1}}{28.94} & {\cellcolor[rgb]{0.925,0.957,1}}\textbf{18.99} 
        & {\cellcolor[rgb]{0.925,0.957,1}}{23.35} & {\cellcolor[rgb]{0.925,0.957,1}}\textbf{4.90} 
        & {\cellcolor[rgb]{0.925,0.957,1}}\textbf{61.04} & {\cellcolor[rgb]{0.925,0.957,1}}\textbf{47.90} 
        & {\cellcolor[rgb]{0.925,0.957,1}}\textbf{44.21} & {\cellcolor[rgb]{0.925,0.957,1}}\textbf{28.21} 
        & {\cellcolor[rgb]{0.925,0.957,1}}\textbf{39.26} & {\cellcolor[rgb]{0.925,0.957,1}}\textbf{26.02} \\
        \bottomrule
    \end{tabular}
    }
\end{table*}

\subsection{Fine-tuning Results}

Tables~\ref{tab:sft-iid-results} and~\ref{tab:sft-ood-results} report the evaluation results of SAC on in-domain and out-of-domain MRQA datasets.

\textbf{Overall Performance.} For in domain results, SAC consistently outperforms all baseline models as shown in Table~\ref{tab:sft-iid-results}. At a 15× compression ratio, compared with the weaker ICAE and the stronger EPL, SAC shows a maximum improvement of 23.5\% F1 / 26.8\% EM and a minimum improvement of 6.7\% F1 / 8.2\% EM for in-domain evaluations. A similar trend can be observed for out-of-domain evaluations in Table~\ref{tab:sft-ood-results} where the maximum improvement is 27.4\% F1 / 28.9\% EM, with a minimum improvement of 6.9\% F1 / 9.2\% EM compared to ICAE and EPL.

\textbf{Comparison with EPL.} As shown in Tables~\ref{tab:sft-iid-results} and~\ref{tab:sft-ood-results}, EPL performs strongly overall thanks to its position layout designed for soft compression. Interestingly, SAC can be seen as building upon EPL~\footnote{This is because the anchor tokens selected by SAC share the same position IDs as the compressed tokens in EPL.} with enhanced compressed token representations introduced in Section~\ref{headings}. The overall results show that the token representation enhancement is effective as SAC outperforms EPL not only overall but also for each individual datasets in MRQA.  



\subsection{Ablation Studies}
In this subsection, we first show ablation studies on two SAC key components, namely anchor embeddings and bidirectional attentions. Then we study the effect of using different strategies to choose anchor tokens. Finally, we show empirically that our carefully designed SAC performs better without autoencoding training, a property that has motivated the SAC design. 

\begin{table*}[ht!]
\small
\centering
    \caption{Component ablation results. We report the average F1/EM performance of the model on in-domain (ID) and out-of-domain (OOD) tasks after removing the bidirectional attention (w/o mask) and the anchor embedding (w/o anchor). Full results on all tasks are provided in the Appendix~\ref{sec:ablation_detailed_res}, including Table~\ref{ablation:iid-full} and Table~\ref{ablation:ood-full}.}
    \label{ablation:component-ablation}
    \resizebox{\textwidth}{!}{
    \begin{tabular}{l|cccccc|cccccc}
        \toprule
        \multirow{4}{*}{\textbf{Methods}} 
        & \multicolumn{6}{c|}{\textbf{In-domain}} 
        & \multicolumn{6}{c}{\textbf{Out-of-domain}} \\
        \cmidrule(r){2-7} 
        \cmidrule(r){8-13}
        & \multicolumn{2}{c}{\textbf{TriviaQA}} 
        & \multicolumn{2}{c}{\textbf{HotpotQA}} 
        & \multicolumn{2}{c|}{\textbf{Average}} 
        & \multicolumn{2}{c}{\textbf{BioASQ}} 
        & \multicolumn{2}{c}{\textbf{TextbookQA}} 
        & \multicolumn{2}{c}{\textbf{Average}} \\
        \cmidrule(r){2-3} \cmidrule(r){4-5} \cmidrule(r){6-7}
        \cmidrule(r){8-9} \cmidrule(r){10-11} \cmidrule(r){12-13}
        & \textbf{F1} & \textbf{EM} 
        & \textbf{F1}  & \textbf{EM} 
        & \textbf{F1}  & \textbf{EM} 
        & \textbf{F1}  & \textbf{EM} 
        & \textbf{F1}  & \textbf{EM} 
        & \textbf{F1}  & \textbf{EM} \\
        \midrule
        {\cellcolor[rgb]{0.925,0.957,1}}\textbf{SAC} 
        & {\cellcolor[rgb]{0.925,0.957,1}}\textbf{65.06} 
        & {\cellcolor[rgb]{0.925,0.957,1}}\textbf{55.93} 
        & {\cellcolor[rgb]{0.925,0.957,1}}\textbf{67.41} 
        & {\cellcolor[rgb]{0.925,0.957,1}}\textbf{50.28} 
        & {\cellcolor[rgb]{0.925,0.957,1}}\textbf{66.24}
        & {\cellcolor[rgb]{0.925,0.957,1}}\textbf{53.11} 
        & {\cellcolor[rgb]{0.925,0.957,1}}\textbf{44.66} 
        & {\cellcolor[rgb]{0.925,0.957,1}}{31.45} 
        & {\cellcolor[rgb]{0.925,0.957,1}}\textbf{52.24} 
        & {\cellcolor[rgb]{0.925,0.957,1}}\textbf{32.93} 
        & {\cellcolor[rgb]{0.925,0.957,1}}\textbf{48.45} 
        & {\cellcolor[rgb]{0.925,0.957,1}}\textbf{32.19}  \\
        SAC(w/o mask) &62.60&53.27 &64.63&47.43 &63.62&50.35 &41.93&30.65 &48.29&29.67 &45.11&30.16 \\
        SAC(w/o anchor) &63.90&54.81 &65.25&48.31 &64.58&51.56 &43.70&\textbf{31.78} &51.59&32.20 &47.65&31.99  \\
        \bottomrule
    \end{tabular}
    }
\end{table*}

\textbf{Component Ablation.} As shown in Table~\ref{ablation:component-ablation}, our ablation study demonstrates the critical roles of the bidirectional attention and anchor embedding. Removing either component results in significant performance degradation in both in-domain (ID) and out-of-domain (OOD) settings. The bidirectional attention mechanism enables anchor tokens to more effectively integrate information from the entire context, producing compressed representations that are more beneficial for downstream tasks. Meanwhile, the anchor embedding provides explicit structural signals that guide the model to accurately identify and process these anchor tokens, thereby ensuring the effectiveness of information compression.

\begin{wraptable}{r}{0.65\linewidth}
\small
\centering
\caption{Token selection results. Different token selection strategies are compared, including Random selection, Lingua-2-based selection~\citep{pan2024llmlingua2datadistillationefficient}, and our uniform selection~\citep{zhao2025positionidsmatterenhanced}. Average F1/EM scores are reported across in-domain (ID) and out-of-domain (OOD) tasks.}
\label{ablation:token-selection}
\renewcommand{\arraystretch}{1.0}
\begin{tabularx}{\linewidth}{l|*{4}{>{\centering\arraybackslash}X}}
\toprule
\multirow{2}{*}{\textbf{Methods}} 
& \multicolumn{2}{c}{\textbf{ID-Average}}
& \multicolumn{2}{c}{\textbf{OOD-Average}} \\
\cmidrule(r){2-3} \cmidrule(r){4-5}
& \textbf{F1} & \textbf{EM} & \textbf{F1} & \textbf{EM} \\
\midrule
\rowcolor[rgb]{0.925,0.957,1} \textbf{SAC} 
& \textbf{63.63} & \textbf{46.95} 
& \cellcolor[rgb]{0.925,0.957,1}\textbf{47.72} 
& \cellcolor[rgb]{0.925,0.957,1}\textbf{32.30} \\
SAC(Random) 
& 59.54 & 43.58 & 45.35 & 30.44 \\
SAC(Lingua-2) 
& 63.26 & 46.54 & 47.43 & 32.13 \\
\bottomrule
\end{tabularx}
\end{wraptable}

\textbf{Token Selection.} 
Table~\ref{ablation:token-selection} shows the results using different token selection strategies for SAC. The results indicate that random selection (\textit{Random}) significantly degrades performance arguably for two reasons: the selected tokens lack semantic importance, and their random positions result in poor global coverage of the context, which together hinder effective representation. In contrast, information-based selection (\textit{Lingua-2}) and our default strategy achieve on par results, and both substantially outperform existing baselines in Tables~\ref{tab:sft-iid-results} and \ref{tab:sft-ood-results}. This demonstrates that the SAC architecture can effectively leverage and enhance any high-quality token selection strategy, rather than relying on a specific choice, highlighting the generality and robustness of the SAC framework.

\begin{table*}[ht!]
    \small
    \centering
    \caption{Ablation study on the effects of autoencoding (AE) and language modeling (LM) objectives.
    }
    \label{ablation:ae-effect}
    \resizebox{\textwidth}{!}{
    \begin{tabular}{l|cccccc|cccccc}
        \toprule
        \multirow{4}{*}{\textbf{Methods}} 
        & \multicolumn{6}{c|}{\textbf{In-domain Average}} 
        & \multicolumn{6}{c}{\textbf{Out-of-domain Average}} \\
        \cmidrule(r){2-7} \cmidrule(r){8-13}
        & \multicolumn{2}{c}{\textbf{5x}} & \multicolumn{2}{c}{\textbf{15x}} & \multicolumn{2}{c|}{\textbf{51x}} 
        & \multicolumn{2}{c}{\textbf{5x}} & \multicolumn{2}{c}{\textbf{15x}} & \multicolumn{2}{c}{\textbf{51x}} \\
        \cmidrule(r){2-3} \cmidrule(r){4-5} \cmidrule(r){6-7}
        \cmidrule(r){8-9} \cmidrule(r){10-11} \cmidrule(r){12-13}
        & \textbf{F1} & \textbf{EM} & \textbf{F1}  & \textbf{EM} & \textbf{F1}  & \textbf{EM} & \textbf{F1}  & \textbf{EM} & \textbf{F1}  & \textbf{EM} & \textbf{F1}  & \textbf{EM} \\
        \midrule
        {\cellcolor[rgb]{0.925,0.957,1}}\textbf{SAC} 
        & {\cellcolor[rgb]{0.925,0.957,1}}\textbf{63.63} 
        & {\cellcolor[rgb]{0.925,0.957,1}}\textbf{46.95} 
        & {\cellcolor[rgb]{0.925,0.957,1}}\textbf{54.95} 
        & {\cellcolor[rgb]{0.925,0.957,1}}\textbf{39.67} 
        & {\cellcolor[rgb]{0.925,0.957,1}}\textbf{46.37}
        & {\cellcolor[rgb]{0.925,0.957,1}}\textbf{33.08} 
        & {\cellcolor[rgb]{0.925,0.957,1}}\textbf{47.72} 
        & {\cellcolor[rgb]{0.925,0.957,1}}\textbf{32.30} 
        & {\cellcolor[rgb]{0.925,0.957,1}}\textbf{39.26} 
        & {\cellcolor[rgb]{0.925,0.957,1}}\textbf{26.02} 
        & {\cellcolor[rgb]{0.925,0.957,1}}\textbf{32.24} 
        & {\cellcolor[rgb]{0.925,0.957,1}}\textbf{21.44}  \\
        500x(w/ LM only) &53.23&38.70 &49.76&35.71 &44.46&31.41  &38.22&25.73 &33.99&22.05 &30.09&18.99  \\
        500x(w/ AE+LM) &55.26&40.14 &49.35&34.98 &43.19&30.26  &38.46&25.40 &34.04&22.09 &30.43&20.09  \\
        SAC(w/ AE only) &56.55&40.34 &49.93&35.33 &43.95&30.64 &42.08&27.98 &35.50&23.29 &28.77&18.32 \\
        SAC(w/ AE+LM) &62.04&45.80 &51.73&36.67 &44.69&31.37 &47.26&32.25 &37.23&23.96 &31.01&19.90 \\
        \bottomrule
    \end{tabular}
    }
\end{table*}

\textbf{AE Effect.} 
Soft compression methods often use AE so that the compressed representations learn to reconstruct the original context. However, AE objective itself can be a limiting factor as the reconstruction target can potentially be misaligned with downstream tasks. SAC is designed with compressed representations directly derived from the contexts to eliminate the need for AE and we show our empirical investigations in Table~\ref{ablation:ae-effect}. 
The experimental results show that training with only the AE objective leads to a substantial performance drop, and even when combined with the LM objective, the performance remains inferior to that of the full SAC model. It is worth noting that while ICAE~\citep{ge2024incontextautoencodercontextcompression} empirically shows that combining AE with LM tasks yields better results, the results do not generalize to other architectures such as 500xCompressor where Table~\ref{ablation:ae-effect} shows a mixed picture for performance. Overall, the results in Table~\ref{ablation:ae-effect} raise questions regarding the necessity of autoencoding tasks and suggest that autoencoding may not be entirely essential for context compression methods.


\begin{table*}[ht!]
\small
\centering
\caption{We compare the performance of SAC and EPL across different model sizes ( \href{https://huggingface.co/meta-llama/Llama-3.2-3B}{Llama-3.2-3B} and \href{https://huggingface.co/meta-llama/Llama-3.1-8B}{Llama-3.1-8B} ) under a fixed 15× compression ratio. Evaluations are conducted in both in-domain (ID) and out-of-domain (OOD) settings, and we report ROUGE-1 F1 and Exact Match (EM) scores.
}
    \label{tab:scalability-results}
    \resizebox{\textwidth}{!}{
    \begin{tabular}{l|cccccc|cccccc}
        \toprule
        \multirow{4}{*}{\textbf{Methods}} 
        & \multicolumn{6}{c|}{\textbf{In-domain}} 
        & \multicolumn{6}{c}{\textbf{Out-of-domain}} \\
        \cmidrule(r){2-7} 
        \cmidrule(r){8-13}
        & \multicolumn{2}{c}{\textbf{SearchQA}} 
        & \multicolumn{2}{c}{\textbf{NQ}} 
        & \multicolumn{2}{c|}{\textbf{Average}} 
        & \multicolumn{2}{c}{\textbf{DROP}} 
        & \multicolumn{2}{c}{\textbf{RE}} 
        & \multicolumn{2}{c}{\textbf{Average}} \\
        \cmidrule(r){2-3} \cmidrule(r){4-5} \cmidrule(r){6-7}
        \cmidrule(r){8-9} \cmidrule(r){10-11} \cmidrule(r){12-13}
        & \textbf{F1} & \textbf{EM} 
        & \textbf{F1}  & \textbf{EM} 
        & \textbf{F1}  & \textbf{EM} 
        & \textbf{F1}  & \textbf{EM} 
        & \textbf{F1}  & \textbf{EM} 
        & \textbf{F1}  & \textbf{EM} \\
        \midrule
        EPL(3B) &{73.61}&{61.56} &{63.59}&{44.43} &{68.60}&{53.00} &{46.20}&{36.46} &{65.73}&{53.70} &{55.97}&{45.08} \\
        {\cellcolor[rgb]{0.925,0.957,1}}\textbf{SAC(3B)} 
        & {\cellcolor[rgb]{0.925,0.957,1}}\textbf{73.88} 
        & {\cellcolor[rgb]{0.925,0.957,1}}\textbf{62.29} 
        & {\cellcolor[rgb]{0.925,0.957,1}}\textbf{65.98} 
        & {\cellcolor[rgb]{0.925,0.957,1}}\textbf{47.18} 
        & {\cellcolor[rgb]{0.925,0.957,1}}\textbf{69.93}
        & {\cellcolor[rgb]{0.925,0.957,1}}\textbf{54.74} 
        & {\cellcolor[rgb]{0.925,0.957,1}}\textbf{48.46} 
        & {\cellcolor[rgb]{0.925,0.957,1}}\textbf{38.39} 
        & {\cellcolor[rgb]{0.925,0.957,1}}\textbf{75.11} 
        & {\cellcolor[rgb]{0.925,0.957,1}}\textbf{62.89} 
        & {\cellcolor[rgb]{0.925,0.957,1}}\textbf{61.76} 
        & {\cellcolor[rgb]{0.925,0.957,1}}\textbf{50.64}  \\
        EPL(8B) &{74.86}&{63.23} &{67.24}&{48.71} &{71.05}&{55.97} &{49.98}&{39.92} &{69.77}&{57.84} &{59.88}&{48.88} \\
        {\cellcolor[rgb]{0.925,0.957,1}}\textbf{SAC(8B)} 
        & {\cellcolor[rgb]{0.925,0.957,1}}\textbf{76.92} 
        & {\cellcolor[rgb]{0.925,0.957,1}}\textbf{65.29} 
        & {\cellcolor[rgb]{0.925,0.957,1}}\textbf{67.77} 
        & {\cellcolor[rgb]{0.925,0.957,1}}\textbf{49.21} 
        & {\cellcolor[rgb]{0.925,0.957,1}}\textbf{72.35}
        & {\cellcolor[rgb]{0.925,0.957,1}}\textbf{57.25} 
        & {\cellcolor[rgb]{0.925,0.957,1}}\textbf{51.55} 
        & {\cellcolor[rgb]{0.925,0.957,1}}\textbf{40.65} 
        & {\cellcolor[rgb]{0.925,0.957,1}}\textbf{78.91} 
        & {\cellcolor[rgb]{0.925,0.957,1}}\textbf{67.71} 
        & {\cellcolor[rgb]{0.925,0.957,1}}\textbf{65.23} 
        & {\cellcolor[rgb]{0.925,0.957,1}}\textbf{54.18}  \\
        \bottomrule
    \end{tabular}
    }
\end{table*}

\subsection{Scalability Results}

To verify the generalizability of SAC across model scales, we conduct additional experiments on the larger Llama-3.2-3B and Llama-3.1-8B models. We show EPL as our baseline which performs the second best in main experiments shown in Table~\ref{tab:sft-iid-results} and ~\ref{tab:sft-ood-results}. As shown in Table~\ref{tab:scalability-results}, SAC outperforms EPL across all model sizes and evaluation settings, consistently achieving higher F1 and EM scores. On the 3B model, SAC achieves an average F1 score of 50.48 and an EM score of 34.73, outperforming the EPL baseline (47.46 / 31.82). On the 8B model, SAC scores 52.31 (F1) and 35.93 (EM), again surpassing the EPL baseline (50.82 / 34.42). We remark that as the model goes larger, the gain does not seem to diminish. These consistent gains demonstrate that SAC effectively generalizes to larger model sizes and maintains its performance advantage. More detailed results on MRQA can be found in Tables~\ref{tab:scalability-iid-results} and~\ref{tab:scalability-ood-results}.

\subsection{Sensitivity Analysis for Compression Ratios}

\begin{wraptable}{r}{0.65\textwidth} 
\small
\centering
\caption{In-domain and out-of-domain performance under different compression ratios.}
\label{tab:sensitivity-analysis-for-compression-ratios}
\renewcommand{\arraystretch}{0.8} 
\begin{tabularx}{\linewidth}{l|*{4}{>{\centering\arraybackslash}X}}
\toprule
\multirow{2}{*}{\textbf{Methods}} 
& \multicolumn{2}{c}{\textbf{In-domain Average}}
& \multicolumn{2}{c}{\textbf{Out-of-domain Average}} \\
\cmidrule(r){2-3} \cmidrule(r){4-5}
& \textbf{F1} & \textbf{EM} & \textbf{F1} & \textbf{EM} \\
\midrule
\multicolumn{5}{c}{{\textit{5x compression}}} \\
\midrule
ICAE &47.53 &33.82 &31.22 &20.51 \\
500x &55.26 &40.14 &38.46 &25.40 \\
EPL  &62.90 &46.33 &46.95 &31.30 \\
\rowcolor[rgb]{0.925,0.957,1} \textbf{SAC} &\textbf{63.63} &\textbf{46.95} &\textbf{47.72} &\textbf{32.30} \\
\midrule
\multicolumn{5}{c}{{\textit{15x compression}}} \\
\midrule
ICAE &44.50 &31.28 &30.81 &20.18 \\
500x &49.35 &34.98 &34.04 &22.09 \\
EPL  &51.52 &36.65 &36.74 &23.83 \\
\rowcolor[rgb]{0.925,0.957,1} \textbf{SAC} &\textbf{54.95} &\textbf{39.67} &\textbf{39.26} &\textbf{26.02} \\
\midrule
\multicolumn{5}{c}{{\textit{51x compression}}} \\
\midrule
ICAE &40.39 &27.99 &27.98 &17.85 \\
500x &43.19 &30.26 &30.43 &20.09 \\
EPL  &43.22 &30.26 &30.22 &19.48 \\
\rowcolor[rgb]{0.925,0.957,1} \textbf{SAC} &\textbf{46.37} &\textbf{33.08} &\textbf{32.24} &\textbf{21.44} \\
\bottomrule
\end{tabularx}
\end{wraptable}
We conduct a detailed evaluation of model performance under different compression ratios (5×, 15×, and 51×), using the same experimental setup as in Table~\ref{tab:sft-iid-results}. The corresponding results are presented in Tables~\ref{tab:sensitivity-analysis-for-compression-ratios}.

As expected, the F1 and EM scores of all methods decrease as the compression ratio increases from 5× to 51×, since higher compression ratios result in more information being discarded. At the highest compression ratio of 51×, the performance of different compression methods varies across datasets (see more detailed results in Appendix~\ref{app:fine-tuning-results}, Tables~\ref{tab:full-sft-iid-results} and~\ref{tab:full-sft-ood-results}): some methods perform well on certain datasets but underperform on others. Nonetheless, SAC consistently achieves the best average performance across all compression ratios, demonstrating strong robustness and stability.

\subsection{Long-Context Summarization Results}
\begin{wraptable}{r}{0.65\textwidth} 
\small
\centering
\caption{We compare the performance of SAC and EPL on long-context summarization tasks, including QMSum~\citep{zhong2021qmsumnewbenchmarkquerybased} and GovReport~\citep{huang2021efficientattentionslongdocument}, reporting ROUGE-1 F1 scores.}
\label{tab:long-context-summarization-results}
\renewcommand{\arraystretch}{1.0} 
\begin{tabularx}{\linewidth}{l|*{3}{>{\centering\arraybackslash}X}}
\toprule
\textbf{Methods} & \textbf{QMSum} & \textbf{GovReport} & \textbf{Average} \\
\midrule
EPL &14.82 &20.40 &17.61  \\
\rowcolor[rgb]{0.925,0.957,1} \textbf{SAC} &\textbf{14.95} &\textbf{22.03} &\textbf{18.49} \\
\bottomrule
\end{tabularx}
\end{wraptable}

To further validate the robustness and usefulness of SAC, we conduct additional experiments on two long-context summarization tasks using the QMSum~\citep{zhong2021qmsumnewbenchmarkquerybased} and GovReport~\citep{huang2021efficientattentionslongdocument} datasets. In these experiments, the models are configured with a maximum input length of 32K tokens and are trained using datasets with a 15$\times$ compression ratio. Both SAC and EPL models are trained and evaluated on the corresponding datasets.

The results in Table~\ref{tab:long-context-summarization-results} show that SAC maintains strong performance on long-context summarization tasks, surpassing EPL across all datasets. Because EPL is trained with AE objective, the results suggest that removing the AE objective for SAC does not impair the model’s capability to model long contexts. Moreover, SAC is able to capture sufficient global information to support tasks that require comprehensive understanding of the entire context, highlighting its adaptability and generalization performance in long-context generation scenarios. More details are provided in Appendix~\ref{app:long-context-summarization-results}.

\section{ANALYSIS}

\subsection{AE Analysis}
\label{subsec:ae_analysis}
We attribute the negative impact of the AE task on SAC primarily to the following two related factors.


\textbf{Intuitive Perspective.} The AE objective requires the compressed representations to reconstruct the complete context, forcing the model to encode a large number of tokens within its limited capacity; some tokens might require significant model memories while not useful for downstream tasks, thereby weakening model's ability to retain task-critical information. 

\textbf{AE-LM Gradient Perspective.} In Figure ~\ref{fig:gradient_cosine_similarity_live}, we visualize the gradient cosine similarity between the autoencoding and the language modeling losses during training. The figure shows that while the two gradients largely align at the very beginning, the cosine similarity quickly approaches zero, suggesting that the two tasks are largely orthogonal in parameter space. This orthogonality causes the optimization of autoencoding to interfere with language modeling updates, thereby hindering downstream task performance.~\footnote{The autoencoding performance will also be impacted by the LM loss for the same orthogonality reason. We show empirically the AE performance in Appendix Table~\ref{table:ae-analysis}.}

\subsection{Attention Visualization}

To understand the unique behavior of compressed models, we analyze the attention patterns of the final layer at a 5$\times$ compression rate. 



At a lower 5× compression rate, as shown in Figure~\ref{fig:5x-attention}, SAC attention map presents a clear positive diagonal, indicating that its compressed tokens primarily attend to local tokens, adhering to the attention prior similar to EPL. In contrast, the attention map of 500xCompressor appears more diffused which arguably complicates learning and impairs its performance. As the compression rate increases (as shown in Figure~\ref{fig:15x-attention} and ~\ref{fig:51x-attention}), EPL attention pattern starts also to diffuse while SAC exhibits a focused attention pattern, with its anchor tokens attending to only a few key original context tokens. Hypothetically, the focused attention pattern retains important token information to handle downstream tasks and explains the superior performance we empirically observe. 

\begin{figure*}[htbp]
    \centering
    \includegraphics[width=\textwidth]{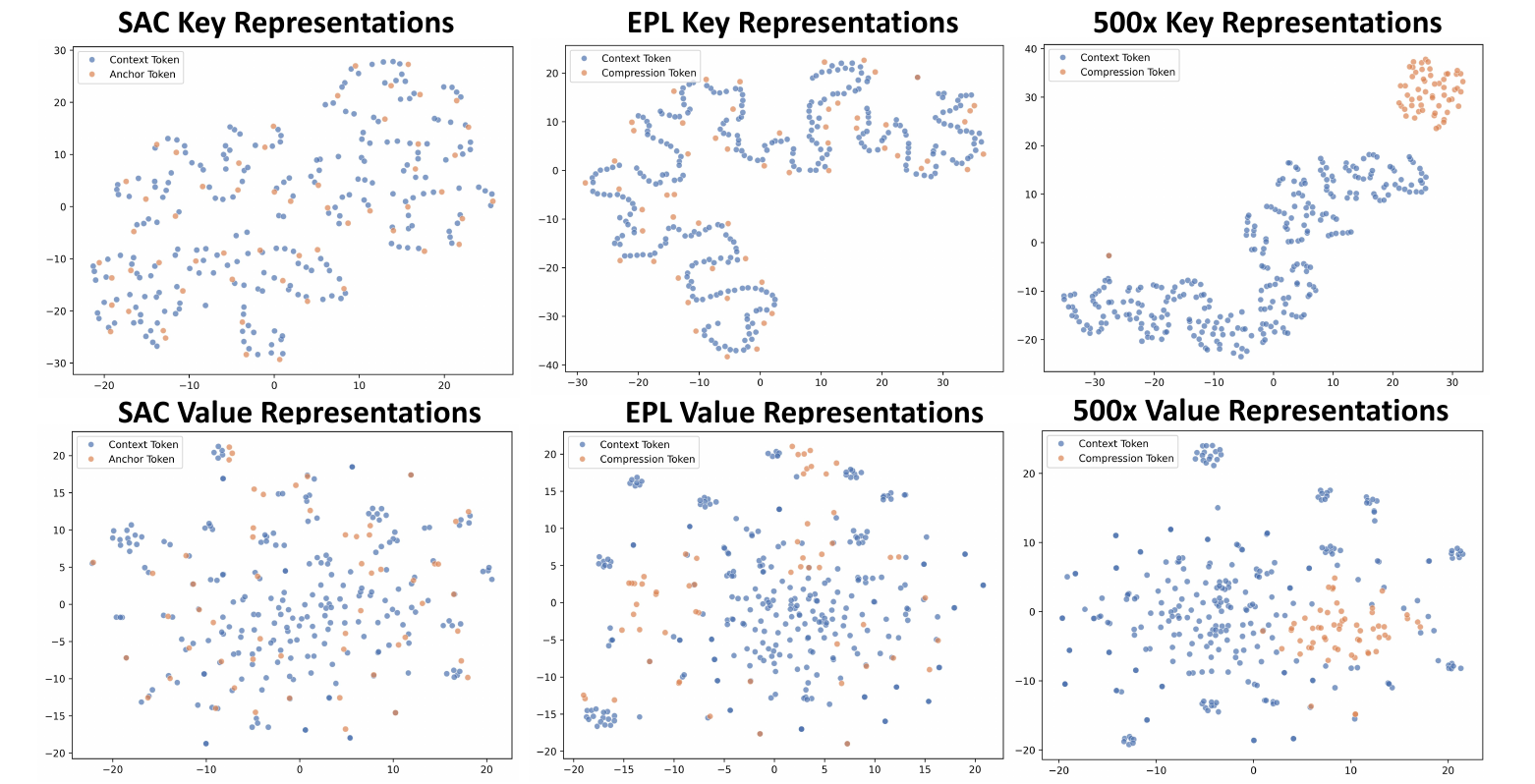}
    \caption{The t-SNE visualization shows the key representations of the final layer KV values for SAC, 500xCompressor~\citep{li2024500xcompressorgeneralizedpromptcompression}, and EPL~\citep{zhao2025positionidsmatterenhanced}, respectively.
}
    \label{fig:tsne-semantic-analysis}
\end{figure*}

\subsection{Representation Analysis}

\textbf{Key Representation Analysis.} In the key representation space (see Figure \ref{fig:tsne-semantic-analysis} up), the compression tokens (orange) of SAC and EPL are distributed relatively close to the context tokens (blue), while the compression tokens of 500xCompressor are clearly separated from the context tokens. This discrepancy arises from the architectural design of each method. EPL modifies the positional IDs of its additional compression tokens to share the same rotational angle (RoPE) as the original context tokens, thereby reducing the distance between them; without such modifications, 500xCompressor compression tokens do not align with context token key representations. In SAC, the anchor tokens are directly embedded within the original context, in consequence their representations maintain close semantic ties with the context tokens. 


\textbf{Value Representation Analysis.} In the value representation space (see Figure \ref{fig:tsne-semantic-analysis} bottom), the anchor tokens of SAC are uniformly distributed across all regions with the value representations of the context tokens, without forming independent sub-clusters. This suggests that SAC's anchor embedding strategy allows for compressed value representations that more closely match the distribution of the original value space. In contrast, although EPL's compression tokens also overlap with the context tokens, they overlap less than SAC's: the compression tokens appear relatively sparse in the core regions and show a slight clustering tendency at the boundaries. This indicates that EPL's value representations still exhibit a degree of semantic shift relative to the original value space, which is even more pronounced in the 500xCompressor.

\section{Conclusion}

In this paper, we propose a novel, autoencoding-free context compression method, \textbf{Semantic-Anchor Compression (SAC)}. Unlike traditional context compression approaches, SAC does not rely on training compression tokens to reconstruct the original input. Instead, it directly selects representative \emph{anchor} tokens from the context and aggregates contextual information into their key-value (KV) representations via a bidirectional attention mechanism. This approach effectively compresses lengthy contexts while avoiding any impairment to the language model's original language modeling capabilities, while traditional context compression can hinder performance by introducing additional compression tokens and the autoencoding tasks, as our analysis shows. Experimental results show that SAC achieves high compression rates while significantly outperforming existing methods across multiple question answering and long-context summarization tasks, demonstrating strong cross-task generalization and an effective balance between compression efficiency and model performance.


\section{Reproducibility statement }
We declare that the work presented in this paper is reproducible. Our data, model, and code have been released at \href{https://github.com/lx-Meteors/SAC}{https://github.com/lx-Meteors/SAC}. This code can be used to reproduce the experimental results. The repository includes detailed instructions for environment setup, running experiments, data processing, and result evaluation.

\section{Acknowledgements}
This work was supported in part by the National Science Foundation of China (Nos. 62276056 and U24A20334), the Yunnan Fundamental Research Projects (No.202401BC070021), the Yunnan Science and Technology Major Project (No. 202502AD080014), the Fundamental Research Funds for the Central Universities (Nos. N25BSS054 and N25BSS094), and the Program of Introducing Talents of Discipline to Universities, Plan 111 (No.B16009).

\bibliography{iclr2026_conference}

@article{huang2025pip,
  title={Pip-kag: Mitigating knowledge conflicts in knowledge-augmented generation via parametric pruning},
  author={Huang, Pengcheng and Liu, Zhenghao and Yan, Yukun and Yi, Xiaoyuan and Chen, Hao and Liu, Zhiyuan and Sun, Maosong and Xiao, Tong and Yu, Ge and Xiong, Chenyan},
  journal={arXiv preprint arXiv:2502.15543},
  year={2025}
}

@article{liu2024forgetting,
  title={Forgetting curve: A reliable method for evaluating memorization capability for long-context models},
  author={Liu, Xinyu and Zhao, Runsong and Huang, Pengcheng and Xiao, Chunyang and Li, Bei and Wang, Jingang and Xiao, Tong and Zhu, Jingbo},
  journal={arXiv preprint arXiv:2410.04727},
  year={2024}
}

@misc{pan2024llmlingua2datadistillationefficient,
      title={LLMLingua-2: Data Distillation for Efficient and Faithful Task-Agnostic Prompt Compression}, 
      author={Zhuoshi Pan and Qianhui Wu and Huiqiang Jiang and Menglin Xia and Xufang Luo and Jue Zhang and Qingwei Lin and Victor Rühle and Yuqing Yang and Chin-Yew Lin and H. Vicky Zhao and Lili Qiu and Dongmei Zhang},
      year={2024},
      eprint={2403.12968},
      archivePrefix={arXiv},
      primaryClass={cs.CL},
      url={https://arxiv.org/abs/2403.12968}, 
}

@inproceedings{li-etal-2023-compressing,
    title = "Compressing Context to Enhance Inference Efficiency of Large Language Models",
    author = "Li, Yucheng  and
      Dong, Bo  and
      Guerin, Frank  and
      Lin, Chenghua",
    editor = "Bouamor, Houda  and
      Pino, Juan  and
      Bali, Kalika",
    booktitle = "Proceedings of the 2023 Conference on Empirical Methods in Natural Language Processing",
    month = dec,
    year = "2023",
    address = "Singapore",
    publisher = "Association for Computational Linguistics",
    url = "https://aclanthology.org/2023.emnlp-main.391/",
    doi = "10.18653/v1/2023.emnlp-main.391",
    pages = "6342--6353"
}

@misc{ge2024incontextautoencodercontextcompression,
      title={In-context Autoencoder for Context Compression in a Large Language Model}, 
      author={Tao Ge and Jing Hu and Lei Wang and Xun Wang and Si-Qing Chen and Furu Wei},
      year={2024},
      eprint={2307.06945},
      archivePrefix={arXiv},
      primaryClass={cs.CL},
      url={https://arxiv.org/abs/2307.06945}, 
}

@inproceedings{li2024500xcompressorgeneralizedpromptcompression,
    title = "500x{C}ompressor: Generalized Prompt Compression for Large Language Models",
    author = "Li, Zongqian  and
      Su, Yixuan  and
      Collier, Nigel",
    editor = "Che, Wanxiang  and
      Nabende, Joyce  and
      Shutova, Ekaterina  and
      Pilehvar, Mohammad Taher",
    booktitle = "Proceedings of the 63rd Annual Meeting of the Association for Computational Linguistics (Volume 1: Long Papers)",
    month = jul,
    year = "2025",
    address = "Vienna, Austria",
    publisher = "Association for Computational Linguistics",
    url = "https://aclanthology.org/2025.acl-long.1219/",
    doi = "10.18653/v1/2025.acl-long.1219",
    pages = "25081--25091",
    ISBN = "979-8-89176-251-0"
}

@misc{zhao2025positionidsmatterenhanced,
      title={Position IDs Matter: An Enhanced Position Layout for Efficient Context Compression in Large Language Models}, 
      author={Runsong Zhao and Xin Liu and Xinyu Liu and Pengcheng Huang and Chunyang Xiao and Tong Xiao and Jingbo Zhu},
      year={2025},
      eprint={2409.14364},
      archivePrefix={arXiv},
      primaryClass={cs.CL},
      url={https://arxiv.org/abs/2409.14364}, 
}

@misc{tang2025gmsaenhancingcontextcompression,
      title={GMSA: Enhancing Context Compression via Group Merging and Layer Semantic Alignment}, 
      author={Jiwei Tang and Zhicheng Zhang and Shunlong Wu and Jingheng Ye and Lichen Bai and Zitai Wang and Tingwei Lu and Jiaqi Chen and Lin Hai and Hai-Tao Zheng and Hong-Gee Kim},
      year={2025},
      eprint={2505.12215},
      archivePrefix={arXiv},
      primaryClass={cs.CL},
      url={https://arxiv.org/abs/2505.12215}, 
}

@misc{zhang2024longcontextcompressionactivation,
      title={Long Context Compression with Activation Beacon}, 
      author={Peitian Zhang and Zheng Liu and Shitao Xiao and Ninglu Shao and Qiwei Ye and Zhicheng Dou},
      year={2024},
      eprint={2401.03462},
      archivePrefix={arXiv},
      primaryClass={cs.CL},
      url={https://arxiv.org/abs/2401.03462}, 
}

@inproceedings{10.5555/3495724.3496517,
author = {Lewis, Patrick and Perez, Ethan and Piktus, Aleksandra and Petroni, Fabio and Karpukhin, Vladimir and Goyal, Naman and K\"{u}ttler, Heinrich and Lewis, Mike and Yih, Wen-tau and Rockt\"{a}schel, Tim and Riedel, Sebastian and Kiela, Douwe},
title = {Retrieval-augmented generation for knowledge-intensive NLP tasks},
year = {2020},
isbn = {9781713829546},
publisher = {Curran Associates Inc.},
address = {Red Hook, NY, USA},
booktitle = {Proceedings of the 34th International Conference on Neural Information Processing Systems},
articleno = {793},
numpages = {16},
location = {Vancouver, BC, Canada},
series = {NIPS '20}
}

@inproceedings{li-etal-2024-loogle,
    title = "{L}oo{GLE}: Can Long-Context Language Models Understand Long Contexts?",
    author = "Li, Jiaqi  and
      Wang, Mengmeng  and
      Zheng, Zilong  and
      Zhang, Muhan",
    editor = "Ku, Lun-Wei  and
      Martins, Andre  and
      Srikumar, Vivek",
    booktitle = "Proceedings of the 62nd Annual Meeting of the Association for Computational Linguistics (Volume 1: Long Papers)",
    month = aug,
    year = "2024",
    address = "Bangkok, Thailand",
    publisher = "Association for Computational Linguistics",
    url = "https://aclanthology.org/2024.acl-long.859/",
    doi = "10.18653/v1/2024.acl-long.859",
    pages = "16304--16333"
}

@article{duan2025chunks,
  title={Chunks as Arms: Multi-Armed Bandit-Guided Sampling for Long-Context LLM Preference Optimization},
  author={Duan, Shaohua and Li, Xinze and Liu, Zhenghao and Yi, Xiaoyuan and Yan, Yukun and Wang, Shuo and Gu, Yu and Yu, Ge and Sun, Maosong},
  journal={arXiv preprint arXiv:2508.13993},
  year={2025}
}

@article{liu-etal-2024-lost,
    title = "Lost in the Middle: How Language Models Use Long Contexts",
    author = "Liu, Nelson F.  and
      Lin, Kevin  and
      Hewitt, John  and
      Paranjape, Ashwin  and
      Bevilacqua, Michele  and
      Petroni, Fabio  and
      Liang, Percy",
    journal = "Transactions of the Association for Computational Linguistics",
    volume = "12",
    year = "2024",
    address = "Cambridge, MA",
    publisher = "MIT Press",
    url = "https://aclanthology.org/2024.tacl-1.9/",
    doi = "10.1162/tacl_a_00638",
    pages = "157--173"
}

@misc{zhang2025surveymultiturninteractioncapabilities,
      title={A Survey on Multi-Turn Interaction Capabilities of Large Language Models}, 
      author={Chen Zhang and Xinyi Dai and Yaxiong Wu and Qu Yang and Yasheng Wang and Ruiming Tang and Yong Liu},
      year={2025},
      eprint={2501.09959},
      archivePrefix={arXiv},
      primaryClass={cs.CL},
      url={https://arxiv.org/abs/2501.09959}, 
}

@misc{yi2025surveyrecentadvancesllmbased,
      title={A Survey on Recent Advances in LLM-Based Multi-turn Dialogue Systems}, 
      author={Zihao Yi and Jiarui Ouyang and Zhe Xu and Yuwen Liu and Tianhao Liao and Haohao Luo and Ying Shen},
      year={2025},
      eprint={2402.18013},
      archivePrefix={arXiv},
      primaryClass={cs.CL},
      url={https://arxiv.org/abs/2402.18013}, 
}

@misc{guan2025evaluatingllmbasedagentsmultiturn,
      title={Evaluating LLM-based Agents for Multi-Turn Conversations: A Survey}, 
      author={Shengyue Guan and Haoyi Xiong and Jindong Wang and Jiang Bian and Bin Zhu and Jian-guang Lou},
      year={2025},
      eprint={2503.22458},
      archivePrefix={arXiv},
      primaryClass={cs.CL},
      url={https://arxiv.org/abs/2503.22458}, 
}

@inproceedings{karpukhin-etal-2020-dense,
    title = "Dense Passage Retrieval for Open-Domain Question Answering",
    author = "Karpukhin, Vladimir  and
      Oguz, Barlas  and
      Min, Sewon  and
      Lewis, Patrick  and
      Wu, Ledell  and
      Edunov, Sergey  and
      Chen, Danqi  and
      Yih, Wen-tau",
    editor = "Webber, Bonnie  and
      Cohn, Trevor  and
      He, Yulan  and
      Liu, Yang",
    booktitle = "Proceedings of the 2020 Conference on Empirical Methods in Natural Language Processing (EMNLP)",
    month = nov,
    year = "2020",
    address = "Online",
    publisher = "Association for Computational Linguistics",
    url = "https://aclanthology.org/2020.emnlp-main.550/",
    doi = "10.18653/v1/2020.emnlp-main.550",
    pages = "6769--6781"
}

@misc{chang2024efficientpromptingmethodslarge,
      title={Efficient Prompting Methods for Large Language Models: A Survey}, 
      author={Kaiyan Chang and Songcheng Xu and Chenglong Wang and Yingfeng Luo and Xiaoqian Liu and Tong Xiao and Jingbo Zhu},
      year={2024},
      eprint={2404.01077},
      archivePrefix={arXiv},
      primaryClass={cs.CL},
      url={https://arxiv.org/abs/2404.01077}, 
}

@inproceedings{li-etal-2025-prompt,
    title = "Prompt Compression for Large Language Models: A Survey",
    author = "Li, Zongqian  and
      Liu, Yinhong  and
      Su, Yixuan  and
      Collier, Nigel",
    editor = "Chiruzzo, Luis  and
      Ritter, Alan  and
      Wang, Lu",
    booktitle = "Proceedings of the 2025 Conference of the Nations of the Americas Chapter of the Association for Computational Linguistics: Human Language Technologies (Volume 1: Long Papers)",
    month = apr,
    year = "2025",
    address = "Albuquerque, New Mexico",
    publisher = "Association for Computational Linguistics",
    url = "https://aclanthology.org/2025.naacl-long.368/",
    doi = "10.18653/v1/2025.naacl-long.368",
    pages = "7182--7195",
    ISBN = "979-8-89176-189-6"
}

@inproceedings{jiang-etal-2023-llmlingua,
    title = "{LLML}ingua: Compressing Prompts for Accelerated Inference of Large Language Models",
    author = "Jiang, Huiqiang  and
      Wu, Qianhui  and
      Lin, Chin-Yew  and
      Yang, Yuqing  and
      Qiu, Lili",
    editor = "Bouamor, Houda  and
      Pino, Juan  and
      Bali, Kalika",
    booktitle = "Proceedings of the 2023 Conference on Empirical Methods in Natural Language Processing",
    month = dec,
    year = "2023",
    address = "Singapore",
    publisher = "Association for Computational Linguistics",
    url = "https://aclanthology.org/2023.emnlp-main.825/",
    doi = "10.18653/v1/2023.emnlp-main.825",
    pages = "13358--13376"
}

@inproceedings{jiang-etal-2024-longllmlingua,
    title = "{L}ong{LLML}ingua: Accelerating and Enhancing {LLM}s in Long Context Scenarios via Prompt Compression",
    author = "Jiang, Huiqiang  and
      Wu, Qianhui  and
      Luo, Xufang  and
      Li, Dongsheng  and
      Lin, Chin-Yew  and
      Yang, Yuqing  and
      Qiu, Lili",
    editor = "Ku, Lun-Wei  and
      Martins, Andre  and
      Srikumar, Vivek",
    booktitle = "Proceedings of the 62nd Annual Meeting of the Association for Computational Linguistics (Volume 1: Long Papers)",
    month = aug,
    year = "2024",
    address = "Bangkok, Thailand",
    publisher = "Association for Computational Linguistics",
    url = "https://aclanthology.org/2024.acl-long.91/",
    doi = "10.18653/v1/2024.acl-long.91",
    pages = "1658--1677"
}

@inproceedings{pan-etal-2024-llmlingua,
    title = "{LLML}ingua-2: Data Distillation for Efficient and Faithful Task-Agnostic Prompt Compression",
    author = {Pan, Zhuoshi  and
      Wu, Qianhui  and
      Jiang, Huiqiang  and
      Xia, Menglin  and
      Luo, Xufang  and
      Zhang, Jue  and
      Lin, Qingwei  and
      R{\"u}hle, Victor  and
      Yang, Yuqing  and
      Lin, Chin-Yew  and
      Zhao, H. Vicky  and
      Qiu, Lili  and
      Zhang, Dongmei},
    editor = "Ku, Lun-Wei  and
      Martins, Andre  and
      Srikumar, Vivek",
    booktitle = "Findings of the Association for Computational Linguistics: ACL 2024",
    month = aug,
    year = "2024",
    address = "Bangkok, Thailand",
    publisher = "Association for Computational Linguistics",
    url = "https://aclanthology.org/2024.findings-acl.57/",
    doi = "10.18653/v1/2024.findings-acl.57",
    pages = "963--981"
}

@misc{wang2020supergluestickierbenchmarkgeneralpurpose,
      title={SuperGLUE: A Stickier Benchmark for General-Purpose Language Understanding Systems}, 
      author={Alex Wang and Yada Pruksachatkun and Nikita Nangia and Amanpreet Singh and Julian Michael and Felix Hill and Omer Levy and Samuel R. Bowman},
      year={2020},
      eprint={1905.00537},
      archivePrefix={arXiv},
      primaryClass={cs.CL},
      url={https://arxiv.org/abs/1905.00537}, 
}

@misc{cerebras2023slimpajama,
  author = {Soboleva, Daria and Al-Khateeb, Faisal and Myers, Robert and Steeves, Jacob R and Hestness, Joel and Dey, Nolan},
  title = {{SlimPajama: A 627B token cleaned and deduplicated version of RedPajama}},
  month = June,
  year = 2023,
  howpublished = {\href{https://www.cerebras.net/blog/slimpajama-a-627b-token-cleaned-and-deduplicated-version-of-redpajama}{https://www.cerebras.net/blog/slimpajama}},
  url = {https://huggingface.co/datasets/cerebras/SlimPajama-627B},
}

@inproceedings{fisch2019mrqa,
    title={{MRQA} 2019 Shared Task: Evaluating Generalization in Reading Comprehension},
    author={Adam Fisch and Alon Talmor and Robin Jia and Minjoon Seo and Eunsol Choi and Danqi Chen},
    booktitle={Proceedings of 2nd Machine Reading for Reading Comprehension (MRQA) Workshop at EMNLP},
    year={2019},
}

@misc{grattafiori2024llama3herdmodels,
      title={The Llama 3 Herd of Models}, 
      author={Aaron Grattafiori and Abhimanyu Dubey},
      year={2024},
      eprint={2407.21783},
      archivePrefix={arXiv},
      primaryClass={cs.AI},
      url={https://arxiv.org/abs/2407.21783}, 
}

@inproceedings{
hu2022lora,
title={Lo{RA}: Low-Rank Adaptation of Large Language Models},
author={Edward J Hu and Yelong Shen and Phillip Wallis and Zeyuan Allen-Zhu and Yuanzhi Li and Shean Wang and Lu Wang and Weizhu Chen},
booktitle={International Conference on Learning Representations},
year={2022},
url={https://openreview.net/forum?id=nZeVKeeFYf9}
}

@inproceedings{lin-2004-rouge,
    title = "{ROUGE}: A Package for Automatic Evaluation of Summaries",
    author = "Lin, Chin-Yew",
    booktitle = "Text Summarization Branches Out",
    month = jul,
    year = "2004",
    address = "Barcelona, Spain",
    publisher = "Association for Computational Linguistics",
    url = "https://aclanthology.org/W04-1013/",
    pages = "74--81"
}

@misc{maalouly2022exactmatchingalgorithmsrelated,
      title={Exact Matching: Algorithms and Related Problems}, 
      author={Nicolas El Maalouly},
      year={2022},
      eprint={2203.13899},
      archivePrefix={arXiv},
      primaryClass={cs.DS},
      url={https://arxiv.org/abs/2203.13899}, 
}

@inproceedings{rajpurkar-etal-2016-squad,
    title = "{SQ}u{AD}: 100,000+ Questions for Machine Comprehension of Text",
    author = "Rajpurkar, Pranav  and
      Zhang, Jian  and
      Lopyrev, Konstantin  and
      Liang, Percy",
    editor = "Su, Jian  and
      Duh, Kevin  and
      Carreras, Xavier",
    booktitle = "Proceedings of the 2016 Conference on Empirical Methods in Natural Language Processing",
    month = nov,
    year = "2016",
    address = "Austin, Texas",
    publisher = "Association for Computational Linguistics",
    url = "https://aclanthology.org/D16-1264/",
    doi = "10.18653/v1/D16-1264",
    pages = "2383--2392"
}

@inproceedings{trischler-etal-2017-newsqa,
    title = "{N}ews{QA}: A Machine Comprehension Dataset",
    author = "Trischler, Adam  and
      Wang, Tong  and
      Yuan, Xingdi  and
      Harris, Justin  and
      Sordoni, Alessandro  and
      Bachman, Philip  and
      Suleman, Kaheer",
    editor = "Blunsom, Phil  and
      Bordes, Antoine  and
      Cho, Kyunghyun  and
      Cohen, Shay  and
      Dyer, Chris  and
      Grefenstette, Edward  and
      Hermann, Karl Moritz  and
      Rimell, Laura  and
      Weston, Jason  and
      Yih, Scott",
    booktitle = "Proceedings of the 2nd Workshop on Representation Learning for {NLP}",
    month = aug,
    year = "2017",
    address = "Vancouver, Canada",
    publisher = "Association for Computational Linguistics",
    url = "https://aclanthology.org/W17-2623/",
    doi = "10.18653/v1/W17-2623",
    pages = "191--200"
}

@inproceedings{joshi-etal-2017-triviaqa,
    title = "{T}rivia{QA}: A Large Scale Distantly Supervised Challenge Dataset for Reading Comprehension",
    author = "Joshi, Mandar  and
      Choi, Eunsol  and
      Weld, Daniel  and
      Zettlemoyer, Luke",
    editor = "Barzilay, Regina  and
      Kan, Min-Yen",
    booktitle = "Proceedings of the 55th Annual Meeting of the Association for Computational Linguistics (Volume 1: Long Papers)",
    month = jul,
    year = "2017",
    address = "Vancouver, Canada",
    publisher = "Association for Computational Linguistics",
    url = "https://aclanthology.org/P17-1147/",
    doi = "10.18653/v1/P17-1147",
    pages = "1601--1611"
}

@article{DBLP:journals/corr/DunnSHGCC17,
  author       = {Matthew Dunn and
                  Levent Sagun and
                  Mike Higgins and
                  V. Ugur G{\"{u}}ney and
                  Volkan Cirik and
                  Kyunghyun Cho},
  title        = {SearchQA: {A} New Q{\&}A Dataset Augmented with Context from a
                  Search Engine},
  journal      = {CoRR},
  volume       = {abs/1704.05179},
  year         = {2017},
  url          = {http://arxiv.org/abs/1704.05179},
  eprinttype    = {arXiv},
  eprint       = {1704.05179},
  bibsource    = {dblp computer science bibliography, https://dblp.org}
}

@inproceedings{yang-etal-2018-hotpotqa,
    title = "{H}otpot{QA}: A Dataset for Diverse, Explainable Multi-hop Question Answering",
    author = "Yang, Zhilin  and
      Qi, Peng  and
      Zhang, Saizheng  and
      Bengio, Yoshua  and
      Cohen, William  and
      Salakhutdinov, Ruslan  and
      Manning, Christopher D.",
    editor = "Riloff, Ellen  and
      Chiang, David  and
      Hockenmaier, Julia  and
      Tsujii, Jun{'}ichi",
    booktitle = "Proceedings of the 2018 Conference on Empirical Methods in Natural Language Processing",
    month = oct # "-" # nov,
    year = "2018",
    address = "Brussels, Belgium",
    publisher = "Association for Computational Linguistics",
    url = "https://aclanthology.org/D18-1259/",
    doi = "10.18653/v1/D18-1259",
    pages = "2369--2380"
}

@article{47761,title	= {Natural Questions: a Benchmark for Question Answering Research},author	= {Tom Kwiatkowski and Jennimaria Palomaki and Olivia Redfield and Michael Collins and Ankur Parikh and Chris Alberti and Danielle Epstein and Illia Polosukhin and Matthew Kelcey and Jacob Devlin and Kenton Lee and Kristina N. Toutanova and Llion Jones and Ming-Wei Chang and Andrew Dai and Jakob Uszkoreit and Quoc Le and Slav Petrov},year	= {2019},journal	= {Transactions of the Association of Computational Linguistics}}

@article{Tsatsaronis2015,
    author = {Tsatsaronis, George and Balikas, Georgios and Malakasiotis, Prodromos and Partalas, Ioannis and Zschunke, Matthias and Alvers, Michael R. and Weissenborn, Dirk and Krithara, Anastasia and Petridis, Sergios and Polychronopoulos, Dimitris and Almirantis, Yannis and Pavlopoulos, John and Baskiotis, Nicolas and Gallinari, Patrick and Arti{\`e}res, Thierry and Ngomo, Axel-Cyrille Ngonga and Heino, Norman and Gaussier, Eric and Barrio-Alvers, Liliana and Schroeder, Michael and Androutsopoulos, Ion and Paliouras, Georgios},
    title = {An overview of the {BIOASQ} large-scale biomedical semantic indexing and question answering competition},
    journal = {BMC Bioinformatics},
    year = {2015},
    volume = {16},
    number = {1},
    pages = {138},
    issn = {1471-2105},
    doi = {10.1186/s12859-015-0564-6},
    url = {https://doi.org/10.1186/s12859-015-0564-6},
    date = {2015-04-30}
}

@inproceedings{dua-etal-2019-drop,
    title = "{DROP}: A Reading Comprehension Benchmark Requiring Discrete Reasoning Over Paragraphs",
    author = "Dua, Dheeru  and
      Wang, Yizhong  and
      Dasigi, Pradeep  and
      Stanovsky, Gabriel  and
      Singh, Sameer  and
      Gardner, Matt",
    editor = "Burstein, Jill  and
      Doran, Christy  and
      Solorio, Thamar",
    booktitle = "Proceedings of the 2019 Conference of the North {A}merican Chapter of the Association for Computational Linguistics: Human Language Technologies, Volume 1 (Long and Short Papers)",
    month = jun,
    year = "2019",
    address = "Minneapolis, Minnesota",
    publisher = "Association for Computational Linguistics",
    url = "https://aclanthology.org/N19-1246/",
    doi = "10.18653/v1/N19-1246",
    pages = "2368--2378"
}

@inproceedings{saha-etal-2018-duorc,
    title = "{D}uo{RC}: Towards Complex Language Understanding with Paraphrased Reading Comprehension",
    author = "Saha, Amrita  and
      Aralikatte, Rahul  and
      Khapra, Mitesh M.  and
      Sankaranarayanan, Karthik",
    editor = "Gurevych, Iryna  and
      Miyao, Yusuke",
    booktitle = "Proceedings of the 56th Annual Meeting of the Association for Computational Linguistics (Volume 1: Long Papers)",
    month = jul,
    year = "2018",
    address = "Melbourne, Australia",
    publisher = "Association for Computational Linguistics",
    url = "https://aclanthology.org/P18-1156/",
    doi = "10.18653/v1/P18-1156",
    pages = "1683--1693"
}

@inproceedings{lai-etal-2017-race,
    title = "{RACE}: Large-scale {R}e{A}ding Comprehension Dataset From Examinations",
    author = "Lai, Guokun  and
      Xie, Qizhe  and
      Liu, Hanxiao  and
      Yang, Yiming  and
      Hovy, Eduard",
    editor = "Palmer, Martha  and
      Hwa, Rebecca  and
      Riedel, Sebastian",
    booktitle = "Proceedings of the 2017 Conference on Empirical Methods in Natural Language Processing",
    month = sep,
    year = "2017",
    address = "Copenhagen, Denmark",
    publisher = "Association for Computational Linguistics",
    url = "https://aclanthology.org/D17-1082/",
    doi = "10.18653/v1/D17-1082",
    pages = "785--794"
}

@inproceedings{levy-etal-2017-zero,
    title = "Zero-Shot Relation Extraction via Reading Comprehension",
    author = "Levy, Omer  and
      Seo, Minjoon  and
      Choi, Eunsol  and
      Zettlemoyer, Luke",
    editor = "Levy, Roger  and
      Specia, Lucia",
    booktitle = "Proceedings of the 21st Conference on Computational Natural Language Learning ({C}o{NLL} 2017)",
    month = aug,
    year = "2017",
    address = "Vancouver, Canada",
    publisher = "Association for Computational Linguistics",
    url = "https://aclanthology.org/K17-1034/",
    doi = "10.18653/v1/K17-1034",
    pages = "333--342"
}

@InProceedings{Kembhavi_2017_CVPR,
author = {Kembhavi, Aniruddha and Seo, Minjoon and Schwenk, Dustin and Choi, Jonghyun and Farhadi, Ali and Hajishirzi, Hannaneh},
title = {Are You Smarter Than a Sixth Grader? Textbook Question Answering for Multimodal Machine Comprehension},
booktitle = {Proceedings of the IEEE Conference on Computer Vision and Pattern Recognition (CVPR)},
month = {July},
year = {2017}
}

@misc{lee2025nvembedimprovedtechniquestraining,
      title={NV-Embed: Improved Techniques for Training LLMs as Generalist Embedding Models}, 
      author={Chankyu Lee and Rajarshi Roy and Mengyao Xu and Jonathan Raiman and Mohammad Shoeybi and Bryan Catanzaro and Wei Ping},
      year={2025},
      eprint={2405.17428},
      archivePrefix={arXiv},
      primaryClass={cs.CL},
      url={https://arxiv.org/abs/2405.17428}, 
}

@misc{behnamghader2024llm2veclargelanguagemodels,
      title={LLM2Vec: Large Language Models Are Secretly Powerful Text Encoders}, 
      author={Parishad BehnamGhader and Vaibhav Adlakha and Marius Mosbach and Dzmitry Bahdanau and Nicolas Chapados and Siva Reddy},
      year={2024},
      eprint={2404.05961},
      archivePrefix={arXiv},
      primaryClass={cs.CL},
      url={https://arxiv.org/abs/2404.05961}, 
}

@inproceedings{wingate-etal-2022-prompt,
    title = "Prompt Compression and Contrastive Conditioning for Controllability and Toxicity Reduction in Language Models",
    author = "Wingate, David  and
      Shoeybi, Mohammad  and
      Sorensen, Taylor",
    editor = "Goldberg, Yoav  and
      Kozareva, Zornitsa  and
      Zhang, Yue",
    booktitle = "Findings of the Association for Computational Linguistics: EMNLP 2022",
    month = dec,
    year = "2022",
    address = "Abu Dhabi, United Arab Emirates",
    publisher = "Association for Computational Linguistics",
    url = "https://aclanthology.org/2022.findings-emnlp.412/",
    doi = "10.18653/v1/2022.findings-emnlp.412",
    pages = "5621--5634"
}

@inproceedings{
mu2023learning,
title={Learning to Compress Prompts with Gist Tokens},
author={Jesse Mu and Xiang Lisa Li and Noah Goodman},
booktitle={Thirty-seventh Conference on Neural Information Processing Systems},
year={2023},
url={https://openreview.net/forum?id=2DtxPCL3T5}
}

@inproceedings{chevalier-etal-2023-adapting,
    title = "Adapting Language Models to Compress Contexts",
    author = "Chevalier, Alexis  and
      Wettig, Alexander  and
      Ajith, Anirudh  and
      Chen, Danqi",
    editor = "Bouamor, Houda  and
      Pino, Juan  and
      Bali, Kalika",
    booktitle = "Proceedings of the 2023 Conference on Empirical Methods in Natural Language Processing",
    month = dec,
    year = "2023",
    address = "Singapore",
    publisher = "Association for Computational Linguistics",
    url = "https://aclanthology.org/2023.emnlp-main.232/",
    doi = "10.18653/v1/2023.emnlp-main.232",
    pages = "3829--3846"
}

@misc{zhong2021qmsumnewbenchmarkquerybased,
      title={QMSum: A New Benchmark for Query-based Multi-domain Meeting Summarization}, 
      author={Ming Zhong and Da Yin and Tao Yu and Ahmad Zaidi and Mutethia Mutuma and Rahul Jha and Ahmed Hassan Awadallah and Asli Celikyilmaz and Yang Liu and Xipeng Qiu and Dragomir Radev},
      year={2021},
      eprint={2104.05938},
      archivePrefix={arXiv},
      primaryClass={cs.CL},
      url={https://arxiv.org/abs/2104.05938}, 
}

@misc{huang2021efficientattentionslongdocument,
      title={Efficient Attentions for Long Document Summarization}, 
      author={Luyang Huang and Shuyang Cao and Nikolaus Parulian and Heng Ji and Lu Wang},
      year={2021},
      eprint={2104.02112},
      archivePrefix={arXiv},
      primaryClass={cs.CL},
      url={https://arxiv.org/abs/2104.02112}, 
}

@misc{chen2025dastcontextawarecompressionllms,
      title={DAST: Context-Aware Compression in LLMs via Dynamic Allocation of Soft Tokens}, 
      author={Shaoshen Chen and Yangning Li and Zishan Xu and Yinghui Li and Xin Su and Zifei Shan and Hai-tao Zheng},
      year={2025},
      eprint={2502.11493},
      archivePrefix={arXiv},
      primaryClass={cs.CL},
      url={https://arxiv.org/abs/2502.11493}, 
}

@misc{wang2024incontextformerlightningfastcompressing,
      title={In-Context Former: Lightning-fast Compressing Context for Large Language Model}, 
      author={Xiangfeng Wang and Zaiyi Chen and Zheyong Xie and Tong Xu and Yongyi He and Enhong Chen},
      year={2024},
      eprint={2406.13618},
      archivePrefix={arXiv},
      primaryClass={cs.CL},
      url={https://arxiv.org/abs/2406.13618}, 
}

@inproceedings{tang-etal-2025-perception,
    title = "Perception Compressor: A Training-Free Prompt Compression Framework in Long Context Scenarios",
    author = "Tang, Jiwei  and
      Xu, Jin  and
      Lu, Tingwei  and
      Zhang, Zhicheng  and
      Zhao, Yiming  and
      Hai, Lin  and
      Zheng, Hai-Tao",
    editor = "Chiruzzo, Luis  and
      Ritter, Alan  and
      Wang, Lu",
    booktitle = "Findings of the Association for Computational Linguistics: NAACL 2025",
    month = apr,
    year = "2025",
    address = "Albuquerque, New Mexico",
    publisher = "Association for Computational Linguistics",
    url = "https://aclanthology.org/2025.findings-naacl.229/",
    doi = "10.18653/v1/2025.findings-naacl.229",
    pages = "4093--4108",
    ISBN = "979-8-89176-195-7"
}

@misc{lv2026datadistributionmattersdatacentric,
      title={Data Distribution Matters: A Data-Centric Perspective on Context Compression for Large Language Model}, 
      author={Kangtao Lv and Jiwei Tang and Langming Liu and Haibin Chen and Weidong Zhang and Shilei Liu and Yongwei Wang and Yujin Yuan and Wenbo Su and Bo Zheng},
      year={2026},
      eprint={2602.01778},
      archivePrefix={arXiv},
      primaryClass={cs.CL},
      url={https://arxiv.org/abs/2602.01778}, 
}

@misc{tang2026readhumancompressingcontext,
      title={Read As Human: Compressing Context via Parallelizable Close Reading and Skimming}, 
      author={Jiwei Tang and Shilei Liu and Zhicheng Zhang and Qingsong Lv and Runsong Zhao and Tingwei Lu and Langming Liu and Haibin Chen and Yujin Yuan and Hai-Tao Zheng and Wenbo Su and Bo Zheng},
      year={2026},
      eprint={2602.01840},
      archivePrefix={arXiv},
      primaryClass={cs.CL},
      url={https://arxiv.org/abs/2602.01840}, 
}

@misc{tang2026comicoarsetofinecontextcompression,
      title={COMI: Coarse-to-fine Context Compression via Marginal Information Gain}, 
      author={Jiwei Tang and Shilei Liu and Zhicheng Zhang and Yujin Yuan and Libin Zheng and Wenbo Su and Bo Zheng},
      year={2026},
      eprint={2602.01719},
      archivePrefix={arXiv},
      primaryClass={cs.CL},
      url={https://arxiv.org/abs/2602.01719}, 
}

@misc{zhao2026cometcollaborativememorytransformer,
      title={CoMeT: Collaborative Memory Transformer for Efficient Long Context Modeling}, 
      author={Runsong Zhao and Shilei Liu and Jiwei Tang and Langming Liu and Haibin Chen and Weidong Zhang and Yujin Yuan and Tong Xiao and Jingbo Zhu and Wenbo Su and Bo Zheng},
      year={2026},
      eprint={2602.01766},
      archivePrefix={arXiv},
      primaryClass={cs.LG},
      url={https://arxiv.org/abs/2602.01766}, 
}
\bibliographystyle{iclr2026_conference}

\newpage

\appendix

\section{Experiment Details}

\label{app:exp}

We perform pretraining and fine-tuning using bf16 precision on 8 NVIDIA RTX 3090 GPUs (24GB). For pretraining, we randomly sample data from the SlimPajama-6B dataset with a token length ranging from 510 to 2040. This data is then split into two halves: one for the auto-encoding (AE) task and the other for the language modeling (LM) task (the AE half is discarded for models without the AE objective). For downstream tasks, we process the MRQA dataset into a (Context, Question, Answer) format for finetuning. Detailed hyperparameters can be found in Table~\ref{tab:hyperparamters}.

\begin{table}[htbp!]
    \centering
    \small
    \caption{Hyperparameters for training}
    \label{tab:hyperparamters}
    \begin{tabular}{ll}
    \toprule
    \textbf{Hyperparameter} & \textbf{Value} \\
    \midrule
    Optimizer              & AdamW \\
    Betas & (0.9, 0.95) \\
    Weight decay & 0.1 \\
    Learning rate          & 1e-4 (pretrain) \\
    & 5e-5 (finetuning) \\
    Scheduler & Constant \\
    Batch size             & 16 \\
    Warmup                 & 300 \\
    Training steps              & 20k (pretrain) \\
    & 20k (finetuning) \\
    Clip norm              & 2.0 \\
    \bottomrule
    \end{tabular}
\end{table}

\section{Detailed Results}

\subsection{Pretraining Results}
As shown in Table~\ref{table:pretrain-results}, our method, SAC, achieves the lowest perplexity (10.79) among all baseline models. This suggests that removing the autoencoding (AE) objective in SAC allows the model to better focus on the language modeling task, thereby improving its predictive capability. Furthermore, since SAC avoids the additional computational overhead from independent compression tokens and the AE task, its training is approximately 31\% faster than ICAE and 26\% faster than 500xCompressor and EPL.

\begin{table*}[ht!]
    \small
    \centering
    \caption{Pretraining comparison of SAC and existing context compression methods, results on LM perplexity and training time.}
    \label{table:pretrain-results}
    \begin{tabular}{l|cc}
        \toprule
        \multirow{1}{*}{\textbf{Methods}} 
        & \multicolumn{1}{c}{\textbf{LM-PPL}}
        & \multicolumn{1}{c}{\textbf{Training Time(hours)}} \\
        
        \midrule
        ICAE &12.35 &3.85  \\
        500x &11.83 &3.60 \\
        EPL &10.88 &3.60 \\
        {\cellcolor[rgb]{0.925,0.957,1}}\textbf{SAC} 
        &{\cellcolor[rgb]{0.925,0.957,1}}\textbf{10.79}
        &{\cellcolor[rgb]{0.925,0.957,1}}\textbf{2.66}  \\
        \bottomrule
    \end{tabular}
\end{table*}

\subsection{Fine-tuning Results}
\label{app:fine-tuning-results}
Tables~\ref{tab:full-sft-iid-results} and~\ref{tab:full-sft-ood-results} report the evaluation results of SAC on in-domain and out-of-domain MRQA datasets, which we analyze from three perspectives: overall performance, effect of compression ratio, and domain generalization.

\textbf{Overall Performance.} SAC consistently outperforms all baselines across a variety of conditions, including compression ratios, and both in-domain and out-of-domain tests, as shown in Tables~\ref{tab:full-sft-iid-results} and \ref{tab:full-sft-ood-results}. Averaging the results of the context compression methods across different compression ratios, SAC shows a maximum improvement of 24.6\% F1 / 28.6\% EM and a minimum improvement of 4.6\% F1 / 5.7\% EM in in-domain evaluations. For out-of-domain tests, the maximum improvement is 32.5\% F1 / 36.2\% EM, with a minimum improvement of 4.6\% F1 / 6.9\% EM.

\textbf{Impact of Compression Ratio.} We conduct a detailed evaluation of model performance under different compression ratios (5×, 15×, and 51×), as shown in Tables~\ref{tab:full-sft-iid-results} and~\ref{tab:full-sft-ood-results}. As expected, F1 and EM scores of all methods decrease with increasing compression ratio, from 5× to 51×, since higher compression ratios result in more information being discarded. At the highest compression rate of 51×, the performance of different compression methods is not consistent. While one method may perform well on certain datasets, it may underperform on others. Nonetheless, SAC consistently achieves the best average performance.

\textbf{Cross-Domain Generalization.} We evaluate the generalization capability of SAC on out-of-domain datasets, as shown in Table~\ref{tab:full-sft-ood-results}. Under all compression ratio constraints, SAC consistently achieves the highest average F1/EM scores among all methods. Specifically, at a 5× compression ratio, SAC attains average F1 and EM scores of 47.72 and 32.30, outperforming the second-best EPL method by 0.77 and 1.0 points, respectively. At a more challenging 15× compression ratio, SAC achieves average F1 and EM scores of 39.26 and 26.02, surpassing EPL by 2.52 and 2.19 points, with an EM improvement approaching 10\%. Even at an extreme 51× compression ratio, SAC maintains average F1 and EM scores of 32.24 and 21.44, still leading EPL by 2.02 and 1.96 points, respectively. These results indicate that the compressed representations learned by SAC exhibit strong cross-domain robustness.

\begin{table*}[ht!]
    \small
    \centering
    \caption{For the finetuning results, we report in-domain performance using ROUGE-1 F1~\citep{lin-2004-rouge} and exact match (EM)~\citep{maalouly2022exactmatchingalgorithmsrelated} scores on the following datasets: SQuAD~\citep{rajpurkar-etal-2016-squad}, NewsQA~\citep{trischler-etal-2017-newsqa}, TriviaQA~\citep{joshi-etal-2017-triviaqa}, SearchQA~\citep{DBLP:journals/corr/DunnSHGCC17}, HotpotQA~\citep{yang-etal-2018-hotpotqa}, and NaturalQuestions (NQ)~\citep{47761}.
}
    \label{tab:full-sft-iid-results}
    \resizebox{\textwidth}{!}{
    \begin{tabular}{l|cccccccccccccc}
        \toprule
        \multirow{2}{*}{\textbf{Methods}} 
        & \multicolumn{2}{c}{\textbf{SQuAD}} 
        & \multicolumn{2}{c}{\textbf{NewsQA}} 
        & \multicolumn{2}{c}{\textbf{TriviaQA}} 
        & \multicolumn{2}{c}{\textbf{SearchQA}}  
        & \multicolumn{2}{c}{\textbf{HotpotQA}} 
        & \multicolumn{2}{c}{\textbf{NQ}}  
        & \multicolumn{2}{c}{\textbf{Average}}\\
        \cmidrule(r){2-3} \cmidrule(r){4-5} \cmidrule(r){6-7} 
        \cmidrule(r){8-9} \cmidrule(r){10-11} \cmidrule(r){12-13} \cmidrule(r){14-15}
        & \textbf{F1} & \textbf{EM} 
        & \textbf{F1} & \textbf{EM} 
        & \textbf{F1} & \textbf{EM} 
        & \textbf{F1} & \textbf{EM} 
        & \textbf{F1} & \textbf{EM} 
        & \textbf{F1} & \textbf{EM} 
        & \textbf{F1} & \textbf{EM}\\ 
        
        \midrule
        Full-FT &77.69&59.71 &63.50&46.04 &68.80&60.54 &73.25&62.07 &74.78&59.26 &71.01&53.47 &71.51&56.85  \\
        Lingua-2 &32.93&19.57 &26.78&13.20 &9.67&8.12 &45.40&31.80 &36.10&22.05 &40.08&22.01 &31.83&19.46  \\
        \midrule
        \multicolumn{14}{c}{{\textit{5x compression constraint}}} \\
        \midrule
        ICAE &36.20&22.12 &28.06&13.77 &54.63&45.59 &65.12&53.06 &48.79&33.40 &52.36&34.99 &47.53&33.82 \\
        500x &51.62&33.63 &39.70&22.63 &57.62&48.76 &66.43&54.38 &59.10&42.20 &57.11&39.26 &55.26&40.14 \\
        EPL &64.72&44.28 &48.74&\textbf{27.45} &63.75&54.54 &69.69&57.73 &67.16&49.79 &63.32&44.16 &62.90&46.33 \\
        {\cellcolor[rgb]{0.925,0.957,1}}\textbf{SAC} 
        & {\cellcolor[rgb]{0.925,0.957,1}}\textbf{65.37} & {\cellcolor[rgb]{0.925,0.957,1}}\textbf{44.83} 
        & {\cellcolor[rgb]{0.925,0.957,1}}\textbf{49.39} & {\cellcolor[rgb]{0.925,0.957,1}}{27.14} 
        & {\cellcolor[rgb]{0.925,0.957,1}}\textbf{65.06} & {\cellcolor[rgb]{0.925,0.957,1}}\textbf{55.93} 
        & {\cellcolor[rgb]{0.925,0.957,1}}\textbf{69.99} & {\cellcolor[rgb]{0.925,0.957,1}}\textbf{58.06} 
        & {\cellcolor[rgb]{0.925,0.957,1}}\textbf{67.41} & {\cellcolor[rgb]{0.925,0.957,1}}\textbf{50.28} 
        & {\cellcolor[rgb]{0.925,0.957,1}}\textbf{64.56} & {\cellcolor[rgb]{0.925,0.957,1}}\textbf{45.44} 
        & {\cellcolor[rgb]{0.925,0.957,1}}\textbf{63.63} & {\cellcolor[rgb]{0.925,0.957,1}}\textbf{46.95} \\
        
        \midrule
        \multicolumn{14}{c}{{\textit{15x compression constraint}}} \\
        \midrule
        ICAE &31.90&18.91 &25.25&11.97 &51.78&42.94 &64.81&52.89 &45.22&30.32 &48.01&30.67 &44.50&31.28 \\
        500x &40.68&24.97 &32.01&16.76 &53.84&44.86 &65.65&53.70 &53.01&36.30 &50.93&33.26 &49.35&34.98 \\
        EPL &44.58&27.91 &33.34&16.69 &56.16&47.09 &66.36&54.13 &54.88&38.38 &53.80&35.71 &51.52&36.65 \\
        {\cellcolor[rgb]{0.925,0.957,1}}\textbf{SAC} 
        & {\cellcolor[rgb]{0.925,0.957,1}}\textbf{47.43} & {\cellcolor[rgb]{0.925,0.957,1}}\textbf{30.25} 
        & {\cellcolor[rgb]{0.925,0.957,1}}\textbf{36.55} & {\cellcolor[rgb]{0.925,0.957,1}}\textbf{18.07} 
        & {\cellcolor[rgb]{0.925,0.957,1}}\textbf{61.13} & {\cellcolor[rgb]{0.925,0.957,1}}\textbf{52.19} 
        & {\cellcolor[rgb]{0.925,0.957,1}}\textbf{68.97} & {\cellcolor[rgb]{0.925,0.957,1}}\textbf{56.76} 
        & {\cellcolor[rgb]{0.925,0.957,1}}\textbf{58.83} & {\cellcolor[rgb]{0.925,0.957,1}}\textbf{41.86} 
        & {\cellcolor[rgb]{0.925,0.957,1}}\textbf{56.79} & {\cellcolor[rgb]{0.925,0.957,1}}\textbf{38.88} 
        & {\cellcolor[rgb]{0.925,0.957,1}}\textbf{54.95} & {\cellcolor[rgb]{0.925,0.957,1}}\textbf{39.67} \\
        
        \midrule
        \multicolumn{14}{c}{{\textit{51x compression constraint}}} \\
        \midrule
        ICAE &26.17&14.58 &22.48&9.69 &47.62&39.23 &64.31&52.80 &38.91&24.78 &42.87&26.86 &40.39&27.99 \\
        500x &30.09&17.11 &25.06&12.20 &50.84&42.13 &64.92&53.29 &42.15&27.32 &46.07&29.53 &43.19&30.26 \\
        EPL &30.09&17.49 &24.49&11.54 &51.15&42.38 &65.12&53.16 &42.19&27.23 &46.29&29.77 &43.22&30.26 \\
        {\cellcolor[rgb]{0.925,0.957,1}}\textbf{SAC} 
        & {\cellcolor[rgb]{0.925,0.957,1}}\textbf{31.81} & {\cellcolor[rgb]{0.925,0.957,1}}\textbf{18.78} 
        & {\cellcolor[rgb]{0.925,0.957,1}}\textbf{27.36} & {\cellcolor[rgb]{0.925,0.957,1}}\textbf{13.56} 
        & {\cellcolor[rgb]{0.925,0.957,1}}\textbf{56.73} & {\cellcolor[rgb]{0.925,0.957,1}}\textbf{47.85} 
        & {\cellcolor[rgb]{0.925,0.957,1}}\textbf{65.82} & {\cellcolor[rgb]{0.925,0.957,1}}\textbf{53.76} 
        & {\cellcolor[rgb]{0.925,0.957,1}}\textbf{48.28} & {\cellcolor[rgb]{0.925,0.957,1}}\textbf{32.84} 
        & {\cellcolor[rgb]{0.925,0.957,1}}\textbf{48.22} & {\cellcolor[rgb]{0.925,0.957,1}}\textbf{31.70} 
        & {\cellcolor[rgb]{0.925,0.957,1}}\textbf{46.37} & {\cellcolor[rgb]{0.925,0.957,1}}\textbf{33.08} \\
        \bottomrule
    \end{tabular}
    }
\end{table*}

\subsection{Ablation Results}

\label{sec:ablation_detailed_res}
In the main text, we discuss the significant performance gains of SAC over all baseline methods. To provide more detailed evidence, we present the full ablation study results here. As shown in Table~\ref{ablation:iid-full} and Table~\ref{ablation:ood-full}, our conclusion holds not only in terms of average performance but is also consistently validated on each individual dataset.

\subsection{Scalability Results}
\label{app:scalability-results}
We use the same data and training settings as described in the paper, with a compression ratio of 15. The training is conducted on 8 A100 GPUs for Llama-3.2-3B and Llama-3.1-8B.

The results are shown in Tables~\ref{tab:scalability-iid-results} and ~\ref{tab:scalability-ood-results}. Compared with the stronger baseline EPL, SAC achieved the best average performance in all cases, demonstrating its ability to effectively scale up to larger model sizes while maintaining its performance advantage.

\subsection{Long-Context Summarization Results}
\label{app:long-context-summarization-results}
To further investigate whether this degradation phenomenon is specific to QA or reflects a broader conflict between AE and downstream tasks, we additionally evaluate SAC without AE and SAC with AE on other long-context tasks. The evaluation metric used is ROUGE-F1.

For the summarization task, we train SAC and EPL on QMSum~\citep{zhong2021qmsumnewbenchmarkquerybased} and GovReport~\citep{huang2021efficientattentionslongdocument} with a maximum input length of 32K tokens. The test results are shown in Table~\ref{table:long-summary-results}. The results indicate that SAC achieves the highest average performance on long-document summarization tasks. Notably, the training data size of QMSum (1.26k) is much smaller than that of GovReport (17.5k), which may explain why SAC(ae+lm) performs slightly better on QMSum, while SAC performs better on the larger GovReport dataset. We hypothesize that as redundant information increases in long contexts and the information capacity of compressed tokens is limited, the AE objective of reconstructing all information may impose additional burden on the model, thereby negatively affecting performance.

We further evaluate on a long-context question answering benchmark with a context length of 24K tokens, using the model checkpoints from Table~\ref{tab:sft-iid-results}, which are trained with a 15× compression ratio on 2K-token contexts. The results are shown in Table~\ref{table:longbench-results}. On both single-document and multi-document question answering tasks, SAC consistently outperforms the best baseline compression methods under a 15× compression ratio. These results indicate that our method can maintain its performance advantages even when applied to longer contexts.

\subsection{FLOPs Results}
We want to clarify a critical advantage of SAC: while SAC does enable bidirectional attention for anchor tokens, it does not require appending additional \textit{k} special tokens to the sequence like 500xCompressor. This means SAC operates on shorter sequences during inference, directly reducing computational overhead.

Let's quantify this advantage. Given a context with shape [\textit{b, s, h}] where \textit{b} is batch size, \textit{s} is sequence length, \textit{h} is hidden size, and \textit{I} is the FFN intermediate size, we compare the theoretical FLOPs:

\begin{table}[htbp]
\centering
\caption{Comparison of FLOPs between SAC and 500xCompressor for different modules.}
\label{tab:flops-comparison}
\begin{tabular}{lcc}
\hline
\textbf{Modules} & \textbf{SAC-FLOPs} & \textbf{500x-FLOPs} \\
\hline
$\mathbf{x(W_Q/W_K/W_V)}$   & $3 \cdot 2 b s h^2$       & $3 \cdot 2 b (s+k) h^2$ \\
$\mathbf{QK^T}$             & $2 b s^2 h$               & $b (s+k)^2 h$           \\
$\mathbf{AV}$               & $2 b s^2 h$               & $b (s+k)^2 h$           \\
$\mathbf{xW_O}$             & $2 b s h^2$               & $2 b (s+k) h^2$         \\
$\mathbf{X_{out} W_{up}}$   & $2 b s h I$               & $2 b (s+k) h I$         \\
$\mathbf{X_{out} W_{gate}}$ & $2 b s h I$               & $2 b (s+k) h I$         \\
$\mathbf{X_{out} W_{down}}$ & $2 b s h I$               & $2 b (s+k) h I$         \\
$\mathbf{sum}$              & $b h s (8 h + 4 s + 6 I)$ & $b h (s+k) [8 h + 2 (s+k) + 6 I]$ \\
\hline
\end{tabular}
\end{table}

Concrete comparison: With typical settings (\textit{b}=1, \textit{s}=510, \textit{h}=2048, \textit{I}=8192):
\begin{itemize}
    \item At 5× compression(\textit{k}=102): 500xCompressor requires \textbf{1.19$\times$ more FLOPs} than SAC
    \item At 10× compression(\textit{k}=51): 500xCompressor requires \textbf{1.08$\times$ more FLOPs} than SAC
\end{itemize}

To validate the theoretical analysis, we additionally measured empirical inference latency. Using a single GPU with batch size 10 over 1000 sequences, we recorded the compression time for both methods. The results are reported in Table \ref{tab:latency}.

\begin{table*}[ht!]
    \small
    \centering
    \caption{Empirical compression time comparison (ms per batch) between SAC and 500xCompressor.}
    \label{tab:latency}
    \begin{tabular}{l|c}
        \toprule
        \multirow{1}{*}{\textbf{Methods}} 
        & \multicolumn{1}{c}{\textbf{Compression Time}} \\
        
        \midrule
        500x &257.43 \\
        {\cellcolor[rgb]{0.925,0.957,1}}\textbf{SAC} 
        &{\cellcolor[rgb]{0.925,0.957,1}}\textbf{243.87} \\
        \bottomrule
    \end{tabular}
\end{table*}

These results show that SAC not only reduces theoretical FLOPs but also achieves lower empirical compression latency. Despite the use of bidirectional attention, SAC is more efficient than 500xCompressor, as the computational cost of bidirectional attention is offset by operating on the original, shorter sequence.

\section{Visualization Analysis}

\subsection{Training Curves Analysis}

\begin{figure}[htbp]
    \centering
    \includegraphics[width=0.99\textwidth]{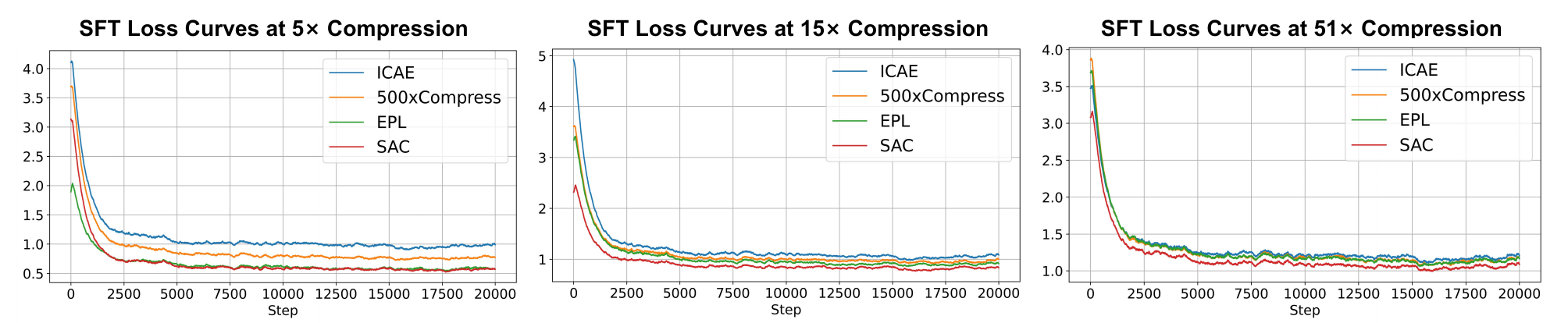}
    \caption{Supervised finetuning loss curves. The figure illustrates the training loss trajectories of different models under three compression ratios: 5×, 15×, and 51×.
}
    \label{fig:SFT-Training-Loss-Curve}
\end{figure}

Figure~\ref{fig:SFT-Training-Loss-Curve} shows the training loss curves at different compression ratios on the MRQA dataset. The training loss of our SAC model consistently converges better than other baseline methods across all compression ratios, which demonstrates that the compressed representations obtained from the SAC architecture are more beneficial for language modeling tasks. Notably, as the compression ratio increases appropriately, the difference in convergence between SAC and the other baselines becomes more significant.

\section{The Use of Large Language Models }
We used a large language model (LLM) as a general-purpose assist tool. The LLM's primary role was in assisting with writing and text editing, such as refining prose and correcting grammar and spelling to ensure the paper's professionalism and fluency. We explicitly state that the LLM was not involved in the core ideation or methodological design of this research. All core contributions of the paper, including the proposal of the methodology, the construction and execution of experiments, and the analysis of results, were performed independently by the authors.

\begin{table*}[ht!]
    \small
    \centering
    \caption{Analysis of autoencoding (AE) reconstruction and language modeling performance under different training objectives.}
    \label{table:ae-analysis}
    \begin{tabular}{l|ccc}
        \toprule
        \multirow{1}{*}{\textbf{Methods}} 
        & \multicolumn{1}{c}{\textbf{AE-PPL}}
        & \multicolumn{1}{c}{\textbf{LM-PPL}}
        & \multicolumn{1}{c}{\textbf{AE-BLEU4}} \\
        
        \midrule
        ICAE &4.08 &12.35 &12.02  \\
        500xCompress &1.09 &11.83 &87.21 \\
        EPL &1.03 &10.88 &97.50 \\
        SAC(ae+lm) &1.03 &11.01 &98.19 \\
        SAC(only ae) &1.00 &- &\textbf{99.94} \\
        {\cellcolor[rgb]{0.925,0.957,1}}\textbf{SAC} 
        &{\cellcolor[rgb]{0.925,0.957,1}}\textbf{-}
        &{\cellcolor[rgb]{0.925,0.957,1}}\textbf{10.79}  
        &{\cellcolor[rgb]{0.925,0.957,1}}\textbf{-} \\
        \bottomrule
    \end{tabular}
\end{table*}

\begin{table*}[ht!]
    \small
    \centering
    \caption{For the finetuning results, we report out-of-domain performance using ROUGE-1 F1 and exact match (EM) scores on the following datasets: BioASQ~\citep{Tsatsaronis2015}, DROP~\citep{dua-etal-2019-drop}, DuoRC~\citep{saha-etal-2018-duorc}, RACE~\citep{lai-etal-2017-race}, Relation Extraction (RE)~\citep{levy-etal-2017-zero}, and TextbookQA (TQA)~\citep{Kembhavi_2017_CVPR}.
    \label{tab:full-sft-ood-results}
}
    \resizebox{\textwidth}{!}{
    \begin{tabular}{l|cccccccccccccc}
        \toprule
        \multirow{2}{*}{\textbf{Methods}} 
        & \multicolumn{2}{c}{\textbf{BioASQ}} 
        & \multicolumn{2}{c}{\textbf{DROP}} 
        & \multicolumn{2}{c}{\textbf{DouRC}} 
        & \multicolumn{2}{c}{\textbf{RACE}}  
        & \multicolumn{2}{c}{\textbf{RE}} 
        & \multicolumn{2}{c}{\textbf{TQA}}  
        & \multicolumn{2}{c}{\textbf{Average}}\\
        \cmidrule(r){2-3} \cmidrule(r){4-5} \cmidrule(r){6-7} 
        \cmidrule(r){8-9} \cmidrule(r){10-11} \cmidrule(r){12-13} \cmidrule(r){14-15}
        & \textbf{F1} & \textbf{EM} 
        & \textbf{F1} & \textbf{EM} 
        & \textbf{F1} & \textbf{EM} 
        & \textbf{F1} & \textbf{EM} 
        & \textbf{F1} & \textbf{EM} 
        & \textbf{F1} & \textbf{EM} 
        & \textbf{F1} & \textbf{EM}\\ 
    
        \midrule
        Full-FT &49.37&36.77 &44.67&34.46 &48.82&35.51 &35.57&9.64 &83.34&72.46 &53.32&32.4 &52.51&36.87  \\
        Lingua-2 &27.76&19.48 &27.28&18.83 &27.07&18.32 &17.54&4.15 &39.30&20.59 &28.42&15.83 &27.90&16.20  \\
        \midrule
        \multicolumn{14}{c}{{\textit{5x compression constraint}}} \\
        \midrule
        ICAE &36.08&26.06 &28.95&21.09 &16.67&10.79 &15.65&3.12 &54.73&41.01 &35.24&20.96 &31.22&20.51 \\
        500x &40.30&28.99 &35.40&25.55 &29.43&19.32 &21.57&4.90 &65.43&50.88 &38.62&22.75 &38.46&25.40\\
        EPL &\textbf{46.05}&\textbf{32.58} &39.94&28.94 &39.10&\textbf{27.12} &\textbf{30.99}&6.08 &76.07&62.31 &49.54&30.74 &46.95&31.30  \\
        {\cellcolor[rgb]{0.925,0.957,1}}\textbf{SAC} 
        & {\cellcolor[rgb]{0.925,0.957,1}}{44.66} & {\cellcolor[rgb]{0.925,0.957,1}}{31.45} 
        & {\cellcolor[rgb]{0.925,0.957,1}}\textbf{41.55} & {\cellcolor[rgb]{0.925,0.957,1}}\textbf{30.87} 
        & {\cellcolor[rgb]{0.925,0.957,1}}\textbf{39.48} & {\cellcolor[rgb]{0.925,0.957,1}}{26.92} 
        & {\cellcolor[rgb]{0.925,0.957,1}}{30.53} & {\cellcolor[rgb]{0.925,0.957,1}}\textbf{6.23} 
        & {\cellcolor[rgb]{0.925,0.957,1}}\textbf{77.87} & {\cellcolor[rgb]{0.925,0.957,1}}\textbf{65.40} 
        & {\cellcolor[rgb]{0.925,0.957,1}}\textbf{52.24} & {\cellcolor[rgb]{0.925,0.957,1}}\textbf{32.93} 
        & {\cellcolor[rgb]{0.925,0.957,1}}\textbf{47.72} & {\cellcolor[rgb]{0.925,0.957,1}}\textbf{32.30} \\
        
        \midrule
        \multicolumn{14}{c}{{\textit{15x compression constraint}}} \\
        \midrule
        ICAE &35.51&24.47 &30.39&21.96 &13.78&9.06 &15.21&3.71 &55.24&40.33 &34.75&21.56 &30.81&20.18 \\
        500x &36.30&25.93 &33.46&23.55 &20.53&12.72 &18.49&3.41 &54.37&41.11 &41.09&25.82 &34.04&22.09 \\
        EPL &40.52&28.52 &32.16&22.29 &25.70&16.39 &20.97&4.01 &59.75&46.34 &41.31&25.42 &36.74&23.83 \\
        {\cellcolor[rgb]{0.925,0.957,1}}\textbf{SAC} 
        & {\cellcolor[rgb]{0.925,0.957,1}}\textbf{41.31} & {\cellcolor[rgb]{0.925,0.957,1}}\textbf{28.66} 
        & {\cellcolor[rgb]{0.925,0.957,1}}\textbf{36.72} & {\cellcolor[rgb]{0.925,0.957,1}}\textbf{27.48} 
        & {\cellcolor[rgb]{0.925,0.957,1}}\textbf{28.94} & {\cellcolor[rgb]{0.925,0.957,1}}\textbf{18.99} 
        & {\cellcolor[rgb]{0.925,0.957,1}}\textbf{23.35} & {\cellcolor[rgb]{0.925,0.957,1}}\textbf{4.90} 
        & {\cellcolor[rgb]{0.925,0.957,1}}\textbf{61.04} & {\cellcolor[rgb]{0.925,0.957,1}}\textbf{47.90} 
        & {\cellcolor[rgb]{0.925,0.957,1}}\textbf{44.21} & {\cellcolor[rgb]{0.925,0.957,1}}\textbf{28.21} 
        & {\cellcolor[rgb]{0.925,0.957,1}}\textbf{39.26} & {\cellcolor[rgb]{0.925,0.957,1}}\textbf{26.02} \\
    
        \midrule
        \multicolumn{14}{c}{{\textit{51x compression constraint}}} \\
        \midrule
        ICAE &33.82&23.67 &27.94&19.29 &11.14&6.86 &14.89&3.41 &47.02&34.02 &33.08&19.83 &27.98&17.85 \\
        500x &32.17&23.07 &\textbf{30.11}&\textbf{21.76} &13.42&8.53 &15.18&2.67 &\textbf{54.62}&\textbf{41.86} &37.10&22.62 &30.43&20.09 \\
        EPL &32.52&22.21 &29.64&20.89 &13.16&8.13 &\textbf{17.15}&3.12 &53.72&40.37 &35.15&22.16 &30.22&19.48 \\
        {\cellcolor[rgb]{0.925,0.957,1}}\textbf{SAC} 
        & {\cellcolor[rgb]{0.925,0.957,1}}\textbf{36.95} & {\cellcolor[rgb]{0.925,0.957,1}}\textbf{26.86} 
        & {\cellcolor[rgb]{0.925,0.957,1}}{29.52} & {\cellcolor[rgb]{0.925,0.957,1}}{20.89} 
        & {\cellcolor[rgb]{0.925,0.957,1}}\textbf{21.85} & {\cellcolor[rgb]{0.925,0.957,1}}\textbf{14.26} 
        & {\cellcolor[rgb]{0.925,0.957,1}}{15.87} & {\cellcolor[rgb]{0.925,0.957,1}}\textbf{4.00} 
        & {\cellcolor[rgb]{0.925,0.957,1}}{48.19} & {\cellcolor[rgb]{0.925,0.957,1}}{36.43} 
        & {\cellcolor[rgb]{0.925,0.957,1}}\textbf{41.05} & {\cellcolor[rgb]{0.925,0.957,1}}\textbf{26.21} 
        & {\cellcolor[rgb]{0.925,0.957,1}}\textbf{32.24} & {\cellcolor[rgb]{0.925,0.957,1}}\textbf{21.44} \\
        \bottomrule
    \end{tabular}
    }
\end{table*}

\begin{table*}[ht!]
    \centering
    \small
    \caption{We conduct ablation studies for SAC under a 5$\times$ compression rate on the in-domain dataset in three sets: component ablation, token selection, and the influence of the autoencoding (AE) task.
}    
    \label{ablation:iid-full}
    \resizebox{\textwidth}{!}{
    \begin{tabular}{l|cccccccccccccc}
        \toprule
        \multirow{2}{*}{\textbf{Methods}} 
        & \multicolumn{2}{c}{\textbf{SQuAD}} 
        & \multicolumn{2}{c}{\textbf{NewsQA}} 
        & \multicolumn{2}{c}{\textbf{TriviaQA}} 
        & \multicolumn{2}{c}{\textbf{SearchQA}}  
        & \multicolumn{2}{c}{\textbf{HotpotQA}} 
        & \multicolumn{2}{c}{\textbf{NQ}}  
        & \multicolumn{2}{c}{\textbf{Average}}\\
        \cmidrule(r){2-3} \cmidrule(r){4-5} \cmidrule(r){6-7} 
        \cmidrule(r){8-9} \cmidrule(r){10-11} 
        \cmidrule(r){12-13} \cmidrule(r){14-15}
        & \textbf{F1} & \textbf{EM} 
        & \textbf{F1} & \textbf{EM} 
        & \textbf{F1} & \textbf{EM} 
        & \textbf{F1} & \textbf{EM} 
        & \textbf{F1} & \textbf{EM} 
        & \textbf{F1} & \textbf{EM} 
        & \textbf{F1} & \textbf{EM}\\ 
    
        \midrule
        \multicolumn{14}{c}{{\textbf{\textit{Component Ablation}}}} \\
        \midrule
        {\cellcolor[rgb]{0.925,0.957,1}}{SAC} 
        & {\cellcolor[rgb]{0.925,0.957,1}}{65.37} 
        & {\cellcolor[rgb]{0.925,0.957,1}}{44.83} 
        & {\cellcolor[rgb]{0.925,0.957,1}}{49.39} 
        & {\cellcolor[rgb]{0.925,0.957,1}}{27.14} 
        & {\cellcolor[rgb]{0.925,0.957,1}}{65.06} 
        & {\cellcolor[rgb]{0.925,0.957,1}}{55.93} 
        & {\cellcolor[rgb]{0.925,0.957,1}}{69.99} 
        & {\cellcolor[rgb]{0.925,0.957,1}}{58.06} 
        & {\cellcolor[rgb]{0.925,0.957,1}}{67.41} 
        & {\cellcolor[rgb]{0.925,0.957,1}}{50.28} 
        & {\cellcolor[rgb]{0.925,0.957,1}}{64.56} 
        & {\cellcolor[rgb]{0.925,0.957,1}}{45.44} 
        & {\cellcolor[rgb]{0.925,0.957,1}}{63.63} 
        & {\cellcolor[rgb]{0.925,0.957,1}}{46.95} \\
        SAC(w/o mask) &60.21&39.93 &45.93&25.74 &62.60&53.27 &66.66&54.82 &64.63&47.43 &61.53&42.55 &60.26&43.96   \\
        SAC(w/o anchor) &61.69&41.72 &46.52&25.45 &63.90&54.81 &68.03&56.17 &65.25&48.31 &62.21&43.88 &61.27&45.06    \\

        \midrule
        \multicolumn{14}{c}{{\textbf{\textit{Token Selection}}}} \\
        \midrule
        SAC(Random) &{58.16}&{39.00} &{42.94}&{22.98} &{62.37}&{53.17} &{69.13}&{57.36} &{63.86}&{46.70} &{60.76}&{42.26} &{59.54}&{43.58}    \\
        SAC(Lingua-2) &64.89&44.28 &48.92&27.11 &64.55&55.13 &69.89&58.04 &67.05&49.74 &64.23&44.93 &63.26&46.54   \\
    
        \midrule
        \multicolumn{14}{c}{{\textbf{\textit{AE Effect}}}} \\
        \midrule
        500x(w/ LM only) &44.71&28.89 &37.24&20.39 &58.97&50.19 &65.67&53.74 &56.74&40.52 &56.07&38.45 &53.23&38.70  \\
        500x(w/ AE+LM) &51.62&33.63 &39.70&22.63 &57.62&48.76 &66.43&54.38 &59.10&42.20 &57.11&39.26 &55.26&40.14  \\
        SAC(w/ AE only) &56.98&37.60 &41.09&20.61 &58.19&49.08 &64.02&51.65 &61.58&44.13 &57.23&38.98 &56.55&40.34   \\
        SAC(w/ AE+LM) &64.68&44.62 &46.64&25.62 &63.34&54.27 &68.40&56.48 &66.61&49.72 &62.56&44.06 &62.04&45.80   \\
        \bottomrule
    \end{tabular}
    }
\end{table*}

\begin{table*}[ht!]
    \centering
    \small
    \caption{We conduct ablation studies for SAC under a 5$\times$ compression rate on the out-of-domain dataset in three sets: component ablation, token selection, and the influence of the autoencoding (AE) task.
}    
    \label{ablation:ood-full}
    \resizebox{\textwidth}{!}{
    \begin{tabular}{l|cccccccccccccc}
        \toprule
        \multirow{2}{*}{\textbf{Methods}} 
        & \multicolumn{2}{c}{\textbf{BioASQ}} 
        & \multicolumn{2}{c}{\textbf{DROP}} 
        & \multicolumn{2}{c}{\textbf{DouRC}} 
        & \multicolumn{2}{c}{\textbf{RACE}}  
        & \multicolumn{2}{c}{\textbf{RE}} 
        & \multicolumn{2}{c}{\textbf{TQA}}  
        & \multicolumn{2}{c}{\textbf{Average}}\\
        \cmidrule(r){2-3} \cmidrule(r){4-5} \cmidrule(r){6-7} 
        \cmidrule(r){8-9} \cmidrule(r){10-11} \cmidrule(r){12-13} \cmidrule(r){14-15}
        & \textbf{F1} & \textbf{EM} 
        & \textbf{F1} & \textbf{EM} 
        & \textbf{F1} & \textbf{EM} 
        & \textbf{F1} & \textbf{EM} 
        & \textbf{F1} & \textbf{EM} 
        & \textbf{F1} & \textbf{EM} 
        & \textbf{F1} & \textbf{EM}\\ 
    
        \midrule
        \multicolumn{14}{c}{{\textbf{\textit{Component Ablation}}}} \\
        \midrule
        {\cellcolor[rgb]{0.925,0.957,1}}{SAC} 
        & {\cellcolor[rgb]{0.925,0.957,1}}{44.66} 
        & {\cellcolor[rgb]{0.925,0.957,1}}{31.45} 
        & {\cellcolor[rgb]{0.925,0.957,1}}{41.55} 
        & {\cellcolor[rgb]{0.925,0.957,1}}{30.87} 
        & {\cellcolor[rgb]{0.925,0.957,1}}{39.48} 
        & {\cellcolor[rgb]{0.925,0.957,1}}{26.92} 
        & {\cellcolor[rgb]{0.925,0.957,1}}{30.53} 
        & {\cellcolor[rgb]{0.925,0.957,1}}{6.23} 
        & {\cellcolor[rgb]{0.925,0.957,1}}{77.87} 
        & {\cellcolor[rgb]{0.925,0.957,1}}{65.40} 
        & {\cellcolor[rgb]{0.925,0.957,1}}{52.24} 
        & {\cellcolor[rgb]{0.925,0.957,1}}{32.93} 
        & {\cellcolor[rgb]{0.925,0.957,1}}{47.72} 
        & {\cellcolor[rgb]{0.925,0.957,1}}{32.30} \\
        SAC(w/o mask) &41.93&30.65 &40.24&28.48 &36.48&23.58 &28.21&5.49 &69.09&55.63 &48.29&29.67 &44.04&28.92 \\
        SAC(w/o anchor) &43.70&31.78 &40.55&30.34 &36.97&25.58 &30.05&6.82 &75.88&62.35 &51.59&32.20 &46.46&31.51  \\

        \midrule
        \multicolumn{14}{c}{{\textbf{\textit{Token Selection}}}} \\
        \midrule
        SAC(Random)  &{45.94}&{34.31} &{40.87}&{30.21} &{35.35}&{24.12} &{29.28}&{5.64} &{70.11}&{56.34} &{50.53}&{32.00} &{45.35}&{30.44} \\
        SAC(Lingua-2) &44.49&31.91 &41.50&29.61 &39.47&26.58 &29.96&7.12 &77.67&65.47 &51.46&32.07 &47.43&32.13 \\
    
        \midrule
        \multicolumn{14}{c}{{\textbf{\textit{AE Effect}}}} \\
        \midrule
        500x(w/ LM only) &43.54&33.11 &35.40&25.82 &27.71&17.59 &19.73&3.86 &62.31&48.27 &40.60&25.75 &38.22&25.73 \\
        500x(w/ AE+LM) &40.30&28.99 &35.40&25.55 &29.43&19.32 &21.57&4.90 &65.43&50.88 &38.62&22.75 &38.46&25.40  \\
        SAC(w/ AE only) &40.85&29.39 &35.32&25.28 &31.55&21.32 &25.86&4.90 &72.29&57.90 &46.61&29.08 &42.08&27.98 \\
        SAC(w/ AE+LM) &44.84&32.31 &41.47&31.14 &39.29&27.58 &30.11&6.23 &77.12&64.42 &50.74&31.87 &47.26&32.26 \\
        \bottomrule
    \end{tabular}
    }
\end{table*}

\begin{table*}[ht!]
    \small
    \centering
    \caption{Experimental Results of the 1B/3B/8B Models on the In-Domain QA Tasks.}
    \label{tab:scalability-iid-results}
    \resizebox{\textwidth}{!}{
    \begin{tabular}{l|cccccccccccccc}
        \toprule
        \multirow{2}{*}{\textbf{Methods}} 
        & \multicolumn{2}{c}{\textbf{SQuAD}} 
        & \multicolumn{2}{c}{\textbf{NewsQA}} 
        & \multicolumn{2}{c}{\textbf{TriviaQA}} 
        & \multicolumn{2}{c}{\textbf{SearchQA}}  
        & \multicolumn{2}{c}{\textbf{HotpotQA}} 
        & \multicolumn{2}{c}{\textbf{NQ}}  
        & \multicolumn{2}{c}{\textbf{Average}}\\
        \cmidrule(r){2-3} \cmidrule(r){4-5} \cmidrule(r){6-7} 
        \cmidrule(r){8-9} \cmidrule(r){10-11} \cmidrule(r){12-13} \cmidrule(r){14-15}
        & \textbf{F1} & \textbf{EM} 
        & \textbf{F1} & \textbf{EM} 
        & \textbf{F1} & \textbf{EM} 
        & \textbf{F1} & \textbf{EM} 
        & \textbf{F1} & \textbf{EM} 
        & \textbf{F1} & \textbf{EM} 
        & \textbf{F1} & \textbf{EM}\\ 
    
        \midrule
        EPL(1B) &{44.58}&{27.91} &{33.34}&{16.69} &{56.16}&{47.09} &{66.36}&{54.13} &{54.88}&{38.38} &{53.80}&{35.71} &{51.52}&{36.65} \\
        {\cellcolor[rgb]{0.925,0.957,1}}\textbf{SAC(1B)} 
        & {\cellcolor[rgb]{0.925,0.957,1}}\textbf{47.43} & {\cellcolor[rgb]{0.925,0.957,1}}\textbf{30.25} 
        & {\cellcolor[rgb]{0.925,0.957,1}}\textbf{36.55} & {\cellcolor[rgb]{0.925,0.957,1}}\textbf{18.07} 
        & {\cellcolor[rgb]{0.925,0.957,1}}\textbf{61.13} & {\cellcolor[rgb]{0.925,0.957,1}}\textbf{52.19} 
        & {\cellcolor[rgb]{0.925,0.957,1}}\textbf{68.97} & {\cellcolor[rgb]{0.925,0.957,1}}\textbf{56.76} 
        & {\cellcolor[rgb]{0.925,0.957,1}}\textbf{58.83} & {\cellcolor[rgb]{0.925,0.957,1}}\textbf{41.86} 
        & {\cellcolor[rgb]{0.925,0.957,1}}\textbf{56.79} & {\cellcolor[rgb]{0.925,0.957,1}}\textbf{38.88} 
        & {\cellcolor[rgb]{0.925,0.957,1}}\textbf{54.95} & {\cellcolor[rgb]{0.925,0.957,1}}\textbf{39.67} \\
        
        EPL(3B) &{62.43}&{42.65} &{46.72}&{26.14} &{68.83}&{59.92} &{73.61}&{61.56} &{68.50}&{51.03} &{63.59}&{44.43} &{63.95}&{47.62} \\
        {\cellcolor[rgb]{0.925,0.957,1}}\textbf{SAC(3B)} 
        & {\cellcolor[rgb]{0.925,0.957,1}}\textbf{63.11} & {\cellcolor[rgb]{0.925,0.957,1}}\textbf{43.51} 
        & {\cellcolor[rgb]{0.925,0.957,1}}\textbf{49.23} & {\cellcolor[rgb]{0.925,0.957,1}}\textbf{28.80} 
        & {\cellcolor[rgb]{0.925,0.957,1}}\textbf{69.65} & {\cellcolor[rgb]{0.925,0.957,1}}\textbf{60.85} 
        & {\cellcolor[rgb]{0.925,0.957,1}}\textbf{73.88} & {\cellcolor[rgb]{0.925,0.957,1}}\textbf{62.29} 
        & {\cellcolor[rgb]{0.925,0.957,1}}\textbf{68.87} & {\cellcolor[rgb]{0.925,0.957,1}}\textbf{51.40} 
        & {\cellcolor[rgb]{0.925,0.957,1}}\textbf{65.98} & {\cellcolor[rgb]{0.925,0.957,1}}\textbf{47.18} 
        & {\cellcolor[rgb]{0.925,0.957,1}}\textbf{65.12} & {\cellcolor[rgb]{0.925,0.957,1}}\textbf{49.01} \\
        
        EPL(8B) &\textbf{65.79}&\textbf{45.55} &\textbf{51.18}&{28.96} &\textbf{72.40}&{63.49} &{74.86}&{63.23} &\textbf{70.98}&{53.65} &{67.24}&{48.71} &{67.08}&{50.60} \\
        {\cellcolor[rgb]{0.925,0.957,1}}\textbf{SAC(8B)} 
        & {\cellcolor[rgb]{0.925,0.957,1}}{64.92} & {\cellcolor[rgb]{0.925,0.957,1}}{44.80} 
        & {\cellcolor[rgb]{0.925,0.957,1}}{49.77} & {\cellcolor[rgb]{0.925,0.957,1}}\textbf{29.20} 
        & {\cellcolor[rgb]{0.925,0.957,1}}{72.37} & {\cellcolor[rgb]{0.925,0.957,1}}\textbf{63.70} 
        & {\cellcolor[rgb]{0.925,0.957,1}}\textbf{76.92} & {\cellcolor[rgb]{0.925,0.957,1}}\textbf{65.29} 
        & {\cellcolor[rgb]{0.925,0.957,1}}{70.80} & {\cellcolor[rgb]{0.925,0.957,1}}\textbf{54.25} 
        & {\cellcolor[rgb]{0.925,0.957,1}}\textbf{67.77} & {\cellcolor[rgb]{0.925,0.957,1}}\textbf{49.21} 
        & {\cellcolor[rgb]{0.925,0.957,1}}\textbf{67.09} & {\cellcolor[rgb]{0.925,0.957,1}}\textbf{51.08} \\
        \bottomrule
    \end{tabular}
    }
\end{table*}

\begin{table*}[ht!]
    \small
    \centering
    \caption{Experimental Results of the 1B/3B/8B Models on the Out-of-Domain QA Tasks.}
    \label{tab:scalability-ood-results}
    \resizebox{\textwidth}{!}{
    \begin{tabular}{l|cccccccccccccc}
        \toprule
        \multirow{2}{*}{\textbf{Methods}} 
        & \multicolumn{2}{c}{\textbf{BioASQ}} 
        & \multicolumn{2}{c}{\textbf{DROP}} 
        & \multicolumn{2}{c}{\textbf{DouRC}} 
        & \multicolumn{2}{c}{\textbf{RACE}}  
        & \multicolumn{2}{c}{\textbf{RE}} 
        & \multicolumn{2}{c}{\textbf{TQA}}  
        & \multicolumn{2}{c}{\textbf{Average}}\\
        \cmidrule(r){2-3} \cmidrule(r){4-5} \cmidrule(r){6-7} 
        \cmidrule(r){8-9} \cmidrule(r){10-11} \cmidrule(r){12-13} \cmidrule(r){14-15}
        & \textbf{F1} & \textbf{EM} 
        & \textbf{F1} & \textbf{EM} 
        & \textbf{F1} & \textbf{EM} 
        & \textbf{F1} & \textbf{EM} 
        & \textbf{F1} & \textbf{EM} 
        & \textbf{F1} & \textbf{EM} 
        & \textbf{F1} & \textbf{EM}\\ 
    
        \midrule
        EPL(1B) &{40.52}&{28.52} &{32.16}&{22.29} &{25.70}&{16.39} &{20.97}&{4.01} &{59.75}&{46.34} &{41.31}&{25.42} &{36.74}&{23.83} \\
        {\cellcolor[rgb]{0.925,0.957,1}}\textbf{SAC(1B)} 
        & {\cellcolor[rgb]{0.925,0.957,1}}\textbf{41.31} & {\cellcolor[rgb]{0.925,0.957,1}}\textbf{28.66} 
        & {\cellcolor[rgb]{0.925,0.957,1}}\textbf{36.72} & {\cellcolor[rgb]{0.925,0.957,1}}\textbf{27.48} 
        & {\cellcolor[rgb]{0.925,0.957,1}}\textbf{28.94} & {\cellcolor[rgb]{0.925,0.957,1}}\textbf{18.99} 
        & {\cellcolor[rgb]{0.925,0.957,1}}\textbf{23.35} & {\cellcolor[rgb]{0.925,0.957,1}}\textbf{4.90} 
        & {\cellcolor[rgb]{0.925,0.957,1}}\textbf{61.04} & {\cellcolor[rgb]{0.925,0.957,1}}\textbf{47.90} 
        & {\cellcolor[rgb]{0.925,0.957,1}}\textbf{44.21} & {\cellcolor[rgb]{0.925,0.957,1}}\textbf{28.21} 
        & {\cellcolor[rgb]{0.925,0.957,1}}\textbf{39.26} & {\cellcolor[rgb]{0.925,0.957,1}}\textbf{26.02} \\
        
        EPL(3B) &{46.76}&{32.98} &{46.20}&{36.46} &{36.81}&{25.32} &{33.80}&{7.57} &{65.73}&{53.70} &{55.46}&{34.86} &{47.46}&{31.82} \\
        {\cellcolor[rgb]{0.925,0.957,1}}\textbf{SAC(3B)} 
        & {\cellcolor[rgb]{0.925,0.957,1}}\textbf{47.96} & {\cellcolor[rgb]{0.925,0.957,1}}\textbf{34.51} 
        & {\cellcolor[rgb]{0.925,0.957,1}}\textbf{48.46} & {\cellcolor[rgb]{0.925,0.957,1}}\textbf{38.39} 
        & {\cellcolor[rgb]{0.925,0.957,1}}\textbf{42.21} & {\cellcolor[rgb]{0.925,0.957,1}}\textbf{29.85} 
        & {\cellcolor[rgb]{0.925,0.957,1}}\textbf{33.61} & {\cellcolor[rgb]{0.925,0.957,1}}\textbf{7.72} 
        & {\cellcolor[rgb]{0.925,0.957,1}}\textbf{75.11} & {\cellcolor[rgb]{0.925,0.957,1}}\textbf{62.89} 
        & {\cellcolor[rgb]{0.925,0.957,1}}\textbf{55.53} & {\cellcolor[rgb]{0.925,0.957,1}}\textbf{35.00} 
        & {\cellcolor[rgb]{0.925,0.957,1}}\textbf{50.48} & {\cellcolor[rgb]{0.925,0.957,1}}\textbf{34.73} \\
        
        EPL(8B) &\textbf{52.52}&\textbf{37.57} &{49.98}&{39.92} &{40.09}&{27.24} &{35.22}&{7.27} &{69.77}&{57.84} &{57.36}&\textbf{36.66} &{50.82}&{34.42} \\
        {\cellcolor[rgb]{0.925,0.957,1}}\textbf{SAC(8B)} 
        & {\cellcolor[rgb]{0.925,0.957,1}}{48.54} & {\cellcolor[rgb]{0.925,0.957,1}}{34.04} 
        & {\cellcolor[rgb]{0.925,0.957,1}}\textbf{51.55} & {\cellcolor[rgb]{0.925,0.957,1}}\textbf{40.65} 
        & {\cellcolor[rgb]{0.925,0.957,1}}\textbf{41.44} & {\cellcolor[rgb]{0.925,0.957,1}}\textbf{29.18} 
        & {\cellcolor[rgb]{0.925,0.957,1}}\textbf{35.52} & {\cellcolor[rgb]{0.925,0.957,1}}\textbf{7.72} 
        & {\cellcolor[rgb]{0.925,0.957,1}}\textbf{78.91} & {\cellcolor[rgb]{0.925,0.957,1}}\textbf{67.71} 
        & {\cellcolor[rgb]{0.925,0.957,1}}\textbf{57.87} & {\cellcolor[rgb]{0.925,0.957,1}}{36.26} 
        & {\cellcolor[rgb]{0.925,0.957,1}}\textbf{52.31} & {\cellcolor[rgb]{0.925,0.957,1}}\textbf{35.93} \\
        \bottomrule
    \end{tabular}
    }
\end{table*}

\begin{table*}[ht!]
    \small
    \centering
    \caption{Model Performance on 32K Long-Context Summarization Tasks.}
    \label{table:long-summary-results}
    \begin{tabular}{l|ccc}
        \toprule
        \multirow{1}{*}{\textbf{Methods}} 
        & \multicolumn{1}{c}{\textbf{QMSum}}
        & \multicolumn{1}{c}{\textbf{GovReport}}
        & \multicolumn{1}{c}{\textbf{Average}} \\
        
        \midrule
        {\cellcolor[rgb]{0.925,0.957,1}}\textbf{SAC} 
        &{\cellcolor[rgb]{0.925,0.957,1}}{14.95}
        &{\cellcolor[rgb]{0.925,0.957,1}}\textbf{22.03}
        &{\cellcolor[rgb]{0.925,0.957,1}}\textbf{18.49}  \\
        SAC(ae+lm) &\textbf{15.32} &20.15 &17.74 \\
        EPL &14.82 &20.40 &17.61 \\
        \bottomrule
    \end{tabular}
\end{table*}

\begin{table}[ht!]
    \small
    \centering
    \caption{Model Performance on 24K Long-Context QA Tasks.}
    \label{table:longbench-results}
    \resizebox{\textwidth}{!}{
    \begin{tabular}{l|cccccccc}
        \toprule
        \multirow{1}{*}{\textbf{Methods}} 
        & \multicolumn{1}{c}{\textbf{MultiFieldQA}}
        & \multicolumn{1}{c}{\textbf{NarrativeQA}}
        & \multicolumn{1}{c}{\textbf{Qasper}}
        & \multicolumn{1}{c}{\textbf{Sigle-Doc-Avg.}}
        & \multicolumn{1}{c}{\textbf{2WikiMQA}}
        & \multicolumn{1}{c}{\textbf{MuSiQue}}
        & \multicolumn{1}{c}{\textbf{HotpotQA}}
        & \multicolumn{1}{c}{\textbf{Multi-Doc-Avg.}} \\
        
        \midrule
        {\cellcolor[rgb]{0.925,0.957,1}}\textbf{SAC} 
        &{\cellcolor[rgb]{0.925,0.957,1}}\textbf{17.75}
        &{\cellcolor[rgb]{0.925,0.957,1}}\textbf{11.33}
        &{\cellcolor[rgb]{0.925,0.957,1}}\textbf{11.67}
        &{\cellcolor[rgb]{0.925,0.957,1}}\textbf{13.58}
        &{\cellcolor[rgb]{0.925,0.957,1}}\textbf{24.78}
        &{\cellcolor[rgb]{0.925,0.957,1}}\textbf{12.36}
        &{\cellcolor[rgb]{0.925,0.957,1}}\textbf{26.03}
        &{\cellcolor[rgb]{0.925,0.957,1}}\textbf{21.06}  \\
        SAC(ae+lm) &{17.38} &{10.49} &{9.05} &{12.31} &{22.47} &{7.76} &{23.88} &{18.04} \\
        EPL &{14.34} &{10.50} &{7.58} &{10.81} &{21.99} &{8.43} &{23.72} &{18.05} \\
        \bottomrule
    \end{tabular}
    }
\end{table}

\begin{figure}[htbp]
    \centering
    \includegraphics[width=0.99\textwidth, height=8cm]{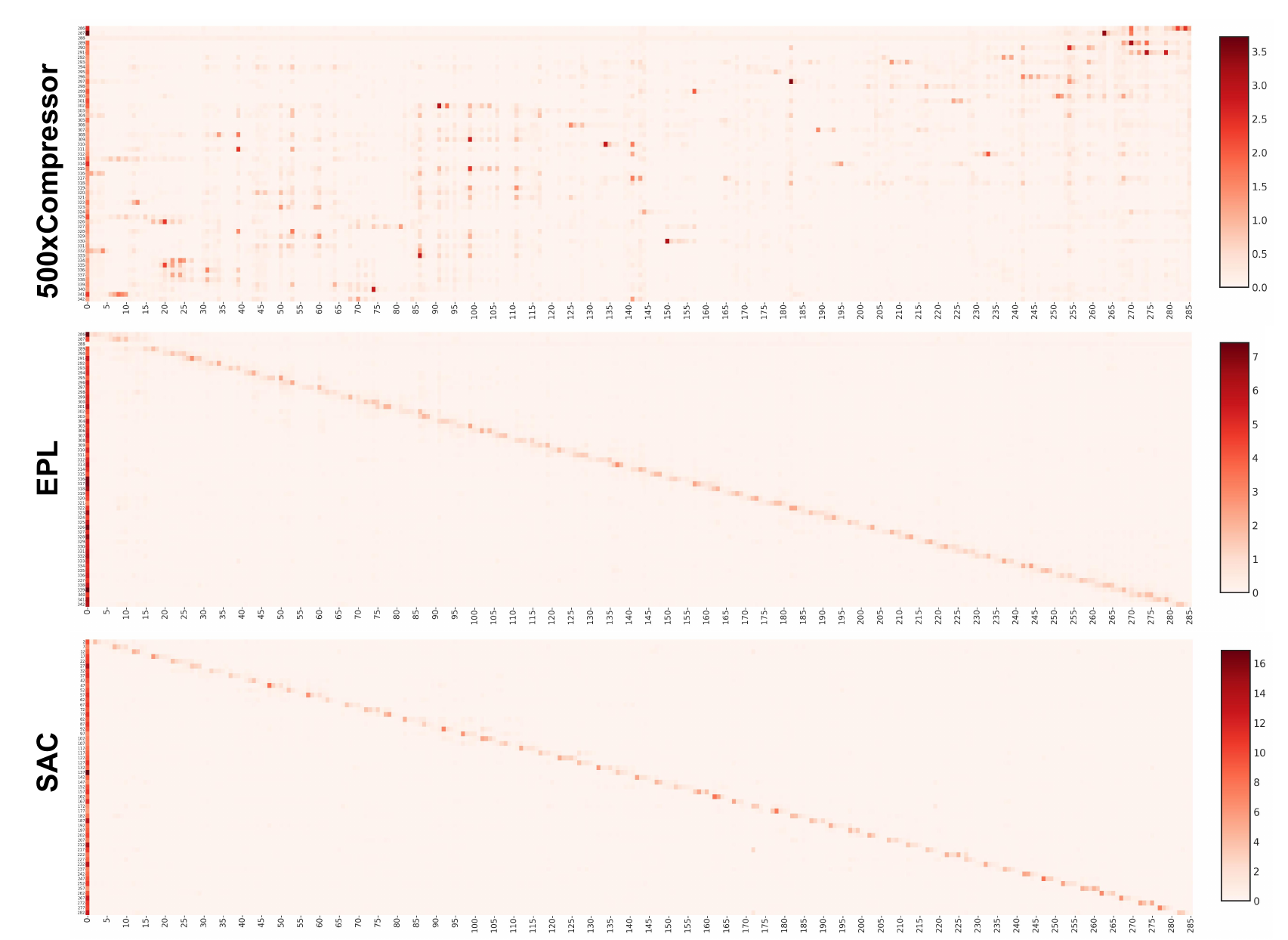}
    \caption{Attention maps of different models finetuned under a 5× compression rate. From top to bottom, the figure displays the final layer attention maps for the 500xCompressor, EPL, and SAC models, respectively. The x-axis represents the original context tokens, and the y-axis represents the compression tokens.}
    \label{fig:5x-attention}
\end{figure}
\begin{figure}[htbp]
    \centering
    \includegraphics[width=0.99\textwidth, height=8cm]{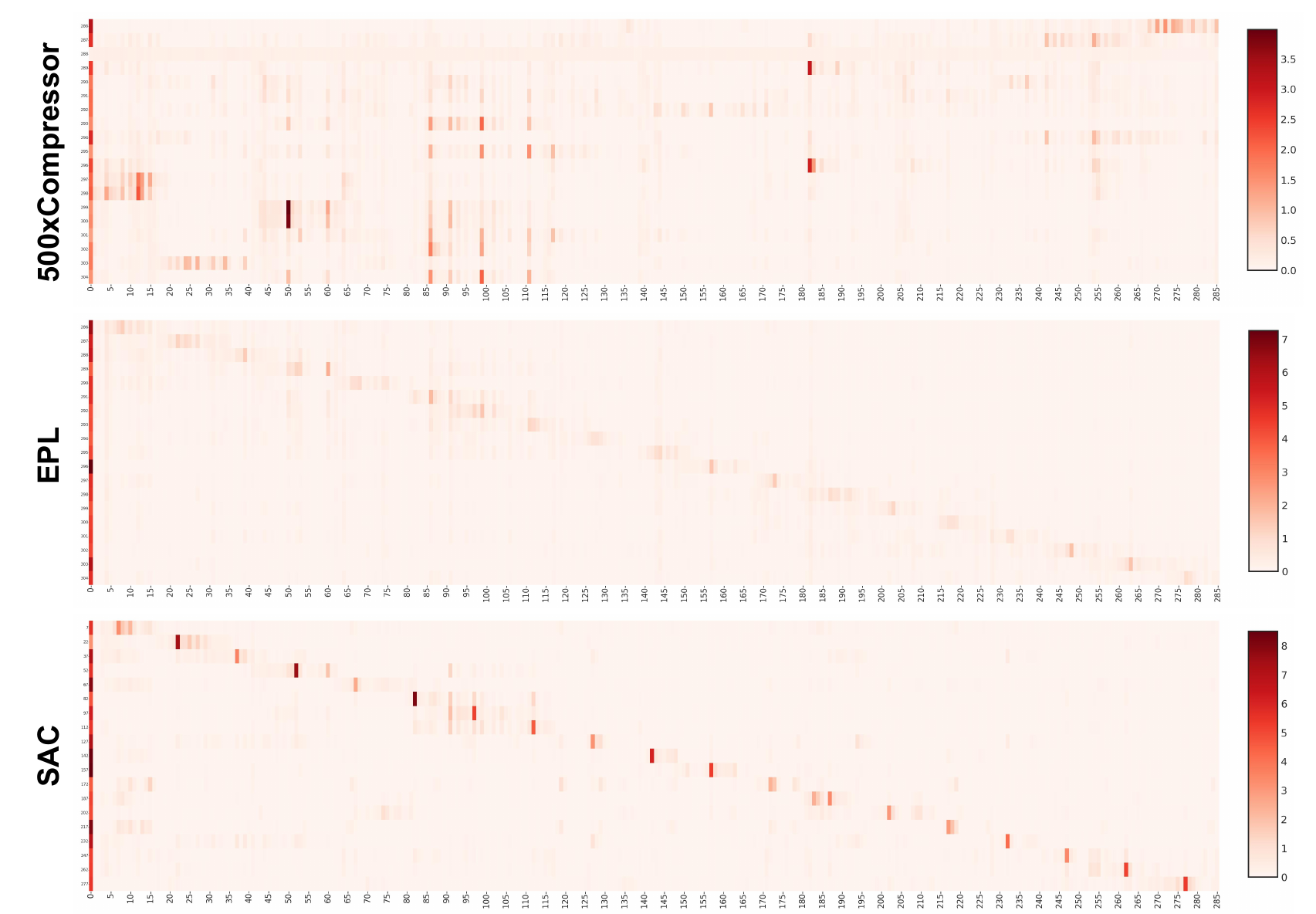}
    \caption{Attention maps of different models finetuned under a 15× compression rate. From top to bottom, the figure displays the final layer attention maps for the 500xCompressor, EPL, and SAC models, respectively. The x-axis represents the original context tokens, and the y-axis represents the compression/anchor tokens.}
    \label{fig:15x-attention}
\end{figure}
\begin{figure}[htbp]
    \centering
    \includegraphics[width=0.99\textwidth, height=8cm]{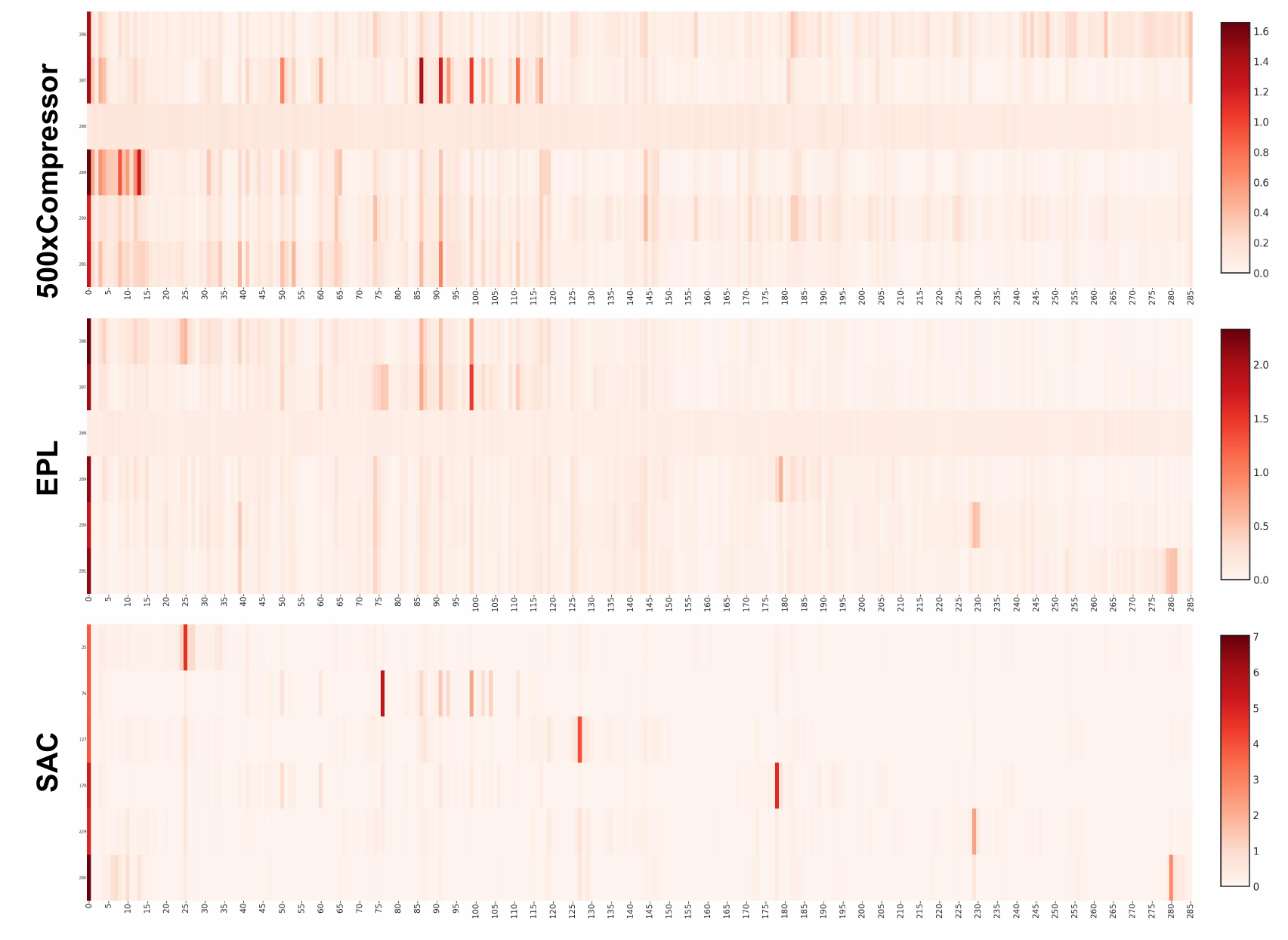}
    \caption{Attention maps of different models finetuned under a 51× compression rate. From top to bottom, the figure displays the final layer attention maps for the 500xCompressor, EPL, and SAC models, respectively. The x-axis represents the original context tokens, and the y-axis represents the compression tokens.}
    \label{fig:51x-attention}
\end{figure}

\begin{figure}[htbp]
    \centering
    \includegraphics[width=0.99\textwidth]{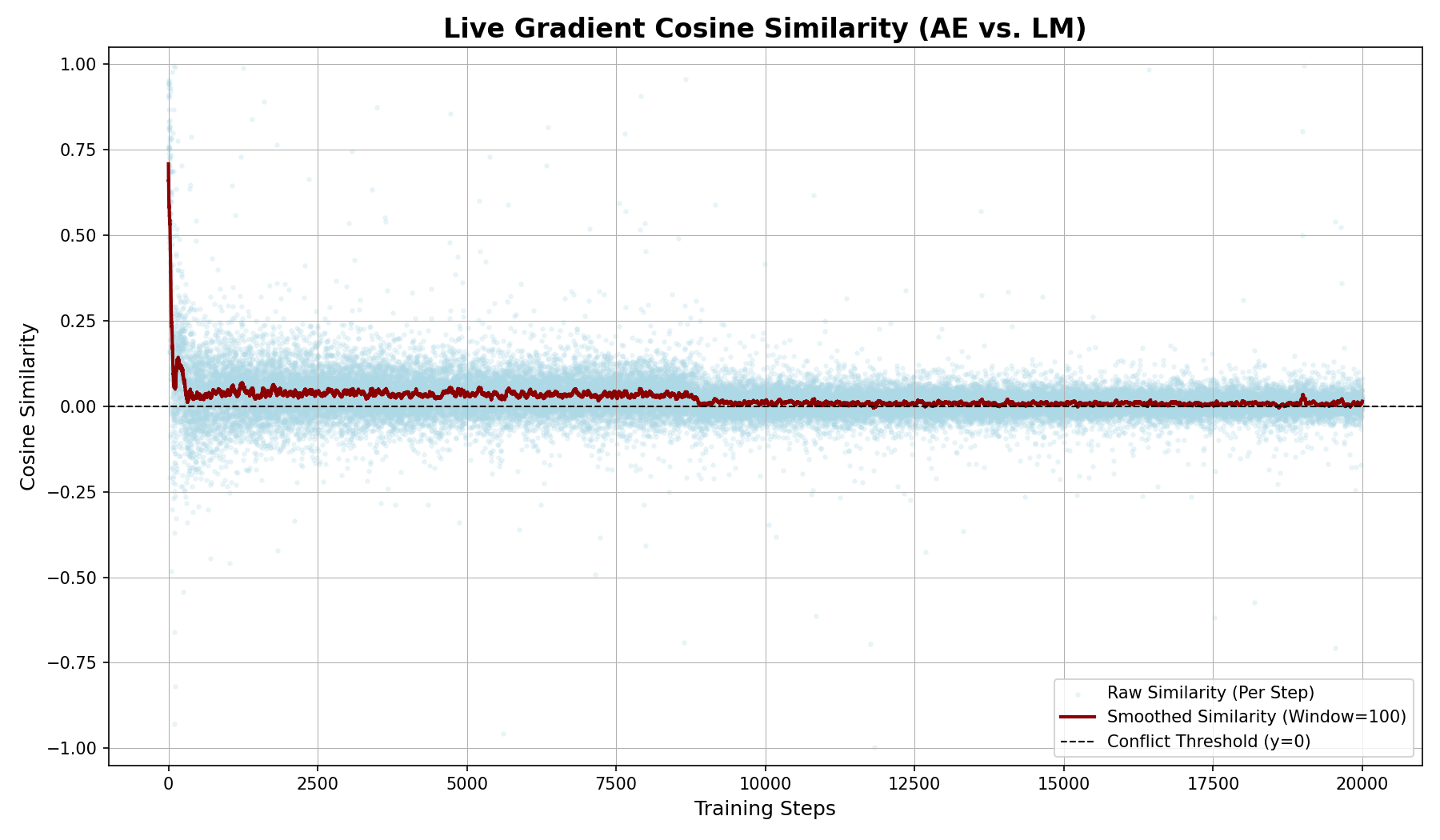}
    \caption{Gradient Cosine Similarity between AutoEncoding (AE) Loss and Language Modeling (LM) Loss.
}
    \label{fig:gradient_cosine_similarity_live}
\end{figure}

\begin{figure}[htbp]
    \centering
    \includegraphics[width=0.99\textwidth]{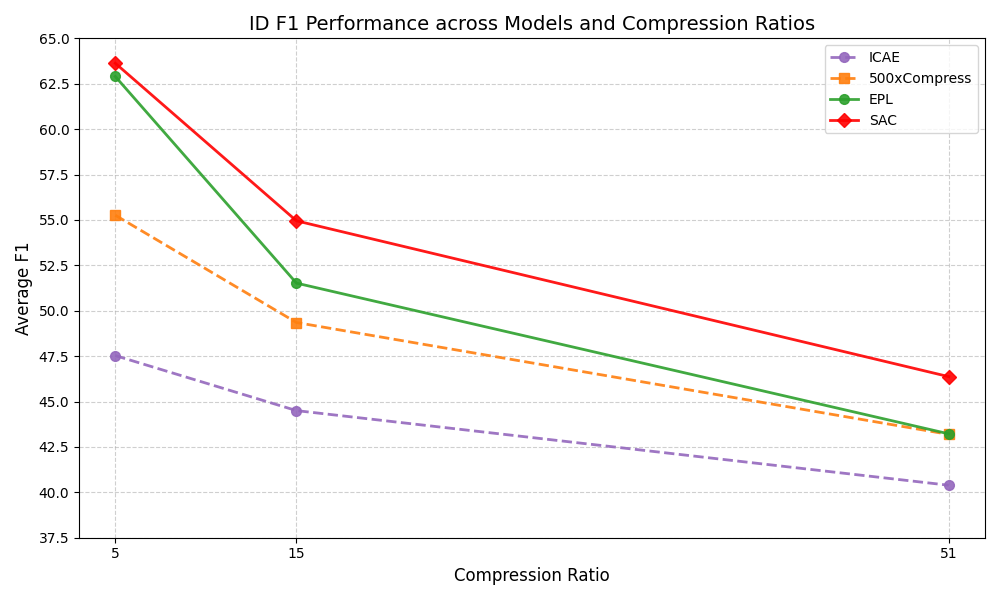}
    \caption{Efficiency and Performance Trade-off Curves on In-Domain (ID) Tasks.
}
    \label{fig:different_compress_ratio_indomain}
\end{figure}


\end{document}